\documentclass[10pt,twocolumn,letterpaper]{article}

\usepackage[pagenumbers]{cvpr} 

\usepackage[dvipsnames]{xcolor}

\definecolor{cvprblue}{rgb}{0.21,0.49,0.74}
\usepackage[pagebackref,breaklinks,colorlinks,citecolor=cvprblue]{hyperref}
\usepackage{algorithmic}
\usepackage[numbers]{natbib}
\usepackage{algorithm}
\usepackage{multirow}
\usepackage{CJK}
\usepackage{enumitem}
\usepackage{indentfirst}
\usepackage{amsmath}

\usepackage{booktabs}
\def\paperID{9532} 
\def\confName{CVPR}
\def\confYear{2024}

\title{Neural-Symbolic VideoQA: Learning Compositional Spatio-Temporal Reasoning for Real-world Video Question Answering}

\author{Lili Liang, Guanglu Sun, Jin Qiu, Lizhong Zhang\\
Harbin University of Science and Technology, School of Computer Science and Technology\\ Harbin, 157000\\
}

\begin{document}
\maketitle
\begin{abstract}
Compositional spatio-temporal reasoning poses a significant challenge in the field of video question answering (VideoQA). Existing approaches struggle to establish effective symbolic reasoning structures, which are crucial for answering compositional spatio-temporal questions. To address this challenge, we propose a neural-symbolic framework called Neural-Symbolic VideoQA (NS-VideoQA), specifically designed for real-world VideoQA tasks. The uniqueness and superiority of NS-VideoQA are two-fold: 1) It proposes a Scene Parser Network (SPN) to transform static-dynamic video scenes into Symbolic Representation (SR), structuralizing persons, objects, relations, and action chronologies. 2) A Symbolic Reasoning Machine (SRM) is designed for top-down question decompositions and bottom-up compositional reasonings. Specifically, a polymorphic program executor is constructed for internally consistent reasoning from SR to the final answer. As a result, Our NS-VideoQA not only improves the compositional spatio-temporal reasoning in real-world VideoQA task, but also enables step-by-step error analysis by tracing the intermediate results. Experimental evaluations on the AGQA Decomp benchmark demonstrate the effectiveness of the proposed NS-VideoQA framework. Empirical studies further confirm that NS-VideoQA exhibits internal consistency in answering compositional questions and significantly improves the capability of spatio-temporal and logical inference for VideoQA tasks.
	
\end{abstract}    
\section{Introduction}
\label{sec:intro}
\begin{figure}[htbp]
	\centering
	\setlength{\belowcaptionskip}{-0.5cm}
	\includegraphics[width=85mm]{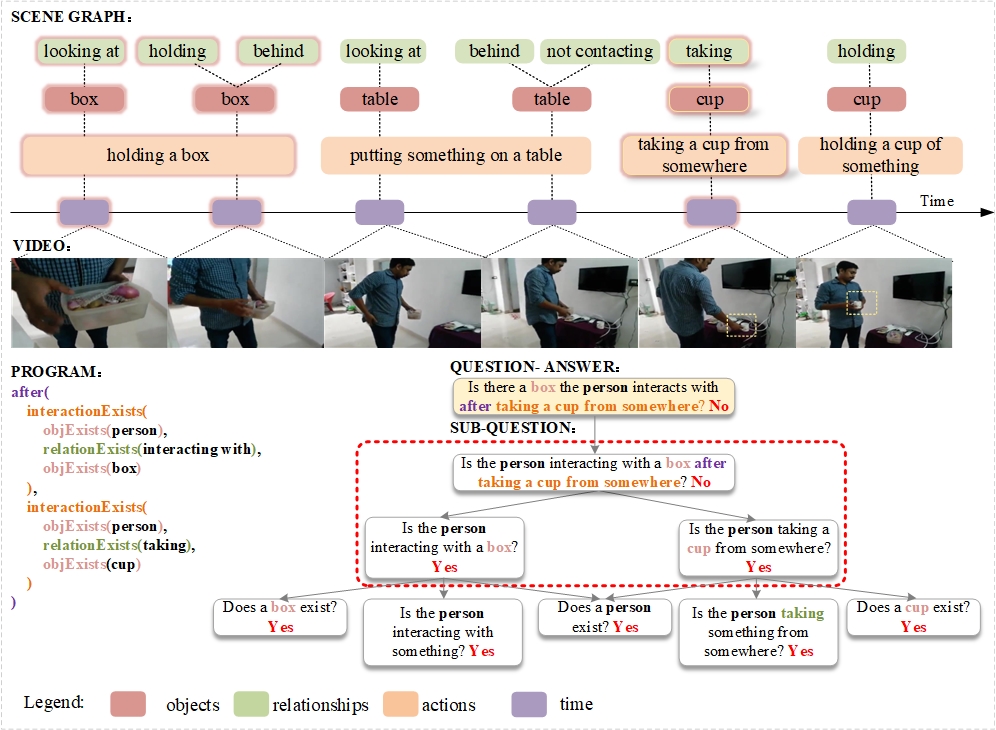}
	\caption{An example of compositional spatio-temporal VideoQA. We mark the objects, relations, actions and time in red, green, orange, and purple respectively. The red dashed box shows a decomposing step from a compositional question to its sub-questions.
	}
	
	\label{demo}
	
\end{figure}
VideoQA in real-world scenarios is an attractive yet challenging task, as it requires multi-modal understanding and reasoning \cite{khurana2021video,sun2021video}. A variety of models have achieved state-of-the-art performance by cross-modal alignment \cite{lee2023dense, lee2022learning,bagad2023test,urooj2023learning} and transformer-based multi-step reasoning \cite{wang2023all,gao2023mist,xiao2022video}. However, these models face challenges when the questions involve complex compositional structures or require understanding spatio-temporal information underlying in videos.

We argue that addressing compositional spatio-temporal questions hinges on two key factors: 1) the cognition of concepts encompassing persons, objects, relations, action chronologies, which contributes to spatio-temporal reasoning; and 2) the compositional reasoning with the analysis of question structures, which effectively mitigates error accumulation during multi-step reasoning.

Using Figure~\ref{demo} as an example, answering the question ``Is there a box the person interacts with after taking a cup from somewhere" requires the ability to recognize the concepts such as the person, objects (box, cup), relations (looking at, taking), and actions (person taking a cup from somewhere) from the given video. Additionally, the question needs to be decomposed into sub-questions, such as ``Is the person interacting with a box?" and ``Is the person taking a cup from somewhere?", which may further require recursive decomposition. Lastly, the ability to reason about the temporal relationship indicated by ``after" using appropriate reasoning rules is crucial. For instance, both sub-questions in Figure~\ref{demo} receive positive answers, but none of the actions after ``taking a cup from somewhere" involve interacting with a box, the final answer would be ``No".

Several approaches have made remarkable progress in understanding and reasoning in visual question answering (VisualQA) \cite{belle2020symbolic,yu2023survey} and VideoQA by incorporating neural-symbolic methods. 
That is, to decouple object attributes with a neural model and to deduce the answer through symbolic reasoning \cite{yi2018neural,yi2019clevrer}. However, these approaches are oriented towards synthesized scenes containing  geometric solids merely, lacks suitability for real-world videos.

To address this limitation, we propose Neural-Symbolic VideoQA (NS-VideoQA), a framework for real-world compositional spatio-temporal VideoQA, designed with neural cognition and symbolic reasoning. For neural cognition, we introduce a Scene Parser Network (SPN) for static and dynamic scenes in videos. SPN transforms non-structural video data into Symbolic Representation (SR), structuralizing persons, objects, relations, and action chronologies. For symbolic reasoning, we proposed Symbolic Reasoning Machine (SRM), where the top-down question decompositions and bottom-up compositional reasonings are conducted. Specifically, we observed that the reasoning rules varies conditioning by the categories of sub-questions. Therefore, a polymorphic program executor is constructed to reason iteratively with internal consistency in various situations. Our NS-VideoQA not only improves the rationality of compositional spatio-temporal reasoning in real-world VideoQA, but also enabled step-by-step error analysis by tracing the intermediate results along the entire reasoning process.

We evaluate our NS-VideoQA framework on the AGQA Decomp dataset \cite{gandhi2022measuring}, which is focused on compositional spatio-temporal questions. We find that our framework outperforms existing purely neural VideoQA models. Further empirical analysis measuring Compositional Accuracy (CA), Right for the Wrong Reasons (RWR) and Internal Consistency (IC) shows that the NS-VideoQA provides superior capability in compositional spatio-temporal reasoning.

In summary, our contributions are:

\begin{itemize}[leftmargin=0.85cm]
	\item[1.]NS-VideoQA is proposed, which is a neural-symbolic framework for VideoQA that enables compositional spatio-temporal reasoning in real-world videos. This is achieved by transforming the video into symbolic representation and conducting iterative reasoning.
	\item[2.]Scene Parser Network, a cognitive model based on Transformer, aims to extract symbolic representations from static-dynamic scenes in videos. This approach converts unstructured videos and questions into structured representations, facilitating symbolic reasoning.
	\item[3.]Symbolic Reasoning Machine, an polymorphic reasoning engine for compositional answer deduction with human-readable inference traces.
\end{itemize}
\section{Related Work}
\label{sec:formatting}
\subsection{Transformer-based VideoQA}
In recent years, predicting answers via visual-textual alignment has been the mainstream methodology towards VideoQA. Typical methods align different modalities via cross-attention \cite{xu2017video}, and conduct multi-step reasoning using multi-hop attention \cite{zhao2018multi,le2019learning} or stacked self-attention layers \cite{li2019beyond}.

Recently, pre-trained Transformers improved both cognition and reasoning in VideoQA. For better cognition, several methods \cite{yang2020bert,urooj2020mmft} integrate tokens from different modalities and pass them through self-attention layers to enhance attention-based cross-modal cognition. However, these methods are insensitive to temporal dependencies required by questions, resulting in a lack of temporal reasoning. To enhance temporal modeling, All in One \cite{wang2023all}, PMT \cite{peng2023efficient}, and RTransformer \cite{zhang2020action} focus on capturing and leveraging temporal information to improve the understanding and reasoning capabilities of the models. For better reasoning, MIST \cite{gao2023mist} and VGT \cite{xiao2022video} utilize different strategies, such as multi-step spatio-temporal reasoning and graph-based reasoning, to capture the underlying structure in video. However, these models struggle when the questions require multi-step reasoning, which reveals the limitation of purely neural models.

Unlike the approaches mentioned above, we utilize a Transformer-based model solely for recognizing symbolic representations, all compositional spatio-temporal reasoning is performed by algorithms based on symbolic logic.
\subsection{Neuro-symbolic Methods}
Neural-symbolic reasoning strives to build a more transparent, precise, and interpretable automated reasoning process by integrating symbolic logic with the powerful perceptual capabilities of neural networks \cite{belle2020symbolic,yu2023survey}. As a pioneering work, IEP \cite{johnson2017inferring} applied neural-symbolic reasoning to VisualQA by generating programs from questions and conducting inference using CNN features. More recently, several work \cite{yi2018neural,mao2018neuro,vedantam2019probabilistic} improve the explainability of IEP-based VisualQA.

VideoQA presents a more challenging task that requires a symbolic reasoning system to model the temporal and dynamic information in videos. CLEVRER \cite{yi2019clevrer}, DCL \cite{chen2020grounding}, and COMPHY \cite{chen2022comphy} are all approaches that aim to enhance symbolic reasoning in VideoQA by employing different strategies, such as task decomposition, concept grounding, and understanding of dynamics and compositionality. However, the models were designed for synthesized videos, not applicable for real-world VideoQA with complex visual scenes.

In this work, we expand the applicability of compositional spatio-temporal VideoQA onto real-world videos.
\subsection{Compositionality Benchmarks}
Recent VideoQA benchmarks \cite{lei2020tvqa+,yu2019activitynet} indicate that state-of-the-art models struggle to answer compositional questions due to multiple error sources. 
AGQA \cite{grunde2021agqa} and its balanced version AGQA 2.0 \cite{grunde2022agqa} was proposed as benchmarks of compositional spatio-temporal reasons for VideoQA. For clear analysis of mispredicting compositional reasoning types, a new benchmark AGQA Decomp \cite{gandhi2022measuring} was constructed with question programs more relevant to video,  providing an systematical environment for compositional reasoning and error analysis. Thus, we adopt AGQA 2.0 to evaluate our NS-VideoQA framework.
\section{NS-VideoQA}

To enhance the ability of neural-symbolic models in answering compositional spatio-temporal questions about real-world videos, we propose the \textbf{Symbolic Representation (SR) of static-dynamic scene} as a symbolic representation of videos. Based on the idea of neural-symbolic reasoning, the NS-VideoQA framework can be divided into two phases.

\begin{figure*}[htbp]
	\centering
	\setlength{\belowcaptionskip}{-0.3cm}
	\includegraphics[width=174mm,height=120mm]{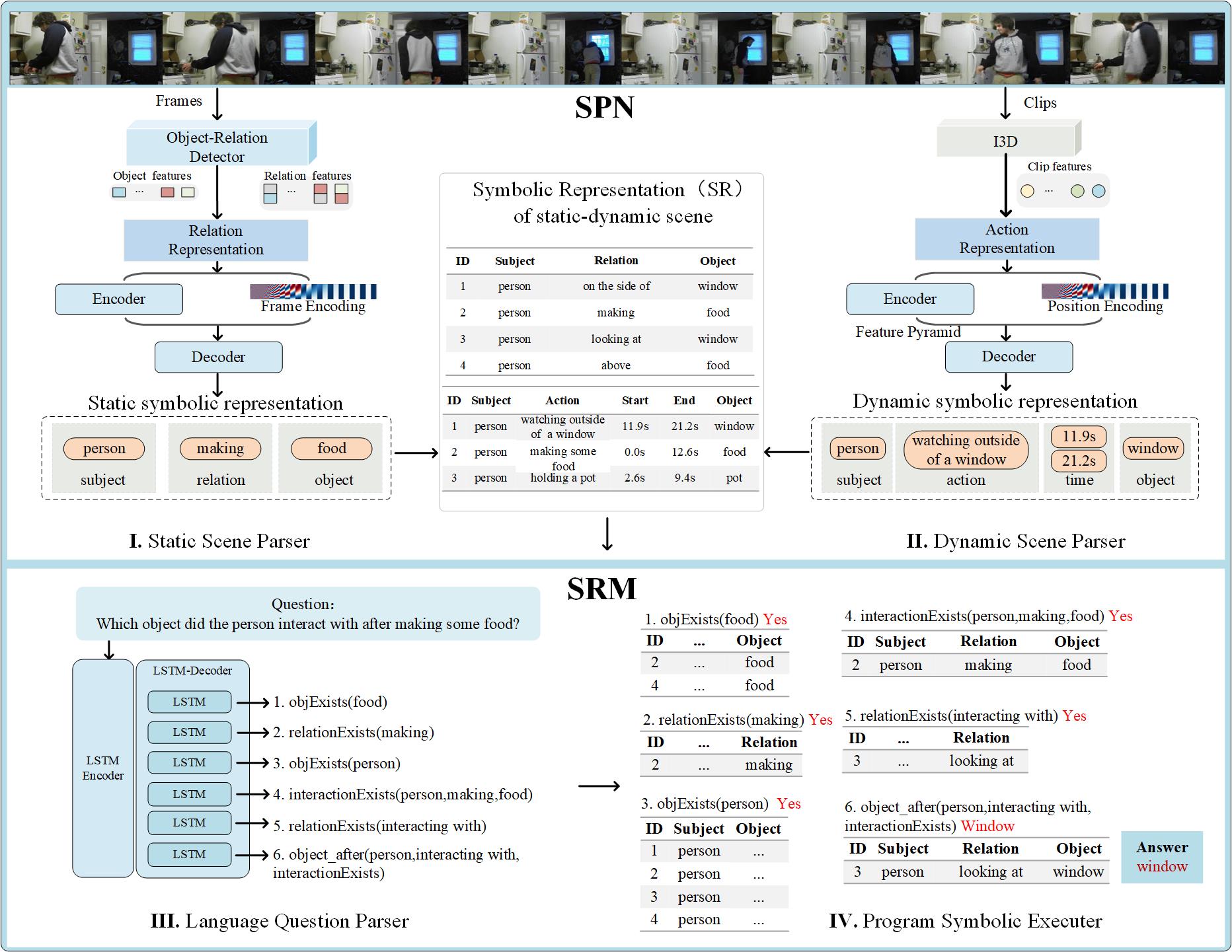}
	\caption{NS-VideoQA uses SPN (I,II) to convert the input video into SR, then uses SRM (III,IV) to decompose the compositional question into a program, and applies reasoning rules iteratively on SR according to the program, finally generates the answer of the compositional question.}
	\label{overview}
\end{figure*}

In the cognition phase, we propose a \textbf{static-dynamic scene parser network (SPN)}. To obtain the SR for a given video, a Static Scene Parser detects persons, objects, and relationships in video frames(Figure~\ref{overview}-I). Concurrently, a Dynamic Scene Parser detects action chronologies in video clips(Figure~\ref{overview}-II). 

In the symbolic reasoning phase, we propose a \textbf{ symbolic reasoning machine (SRM)}. To generate the answer for a given question, a Language Question Parser transforms the question from natural language into a symbolic program (Figure~\ref{overview}-III). Subsequently, a Program Symbolic Executor performs the program based on SR (Figure~\ref{overview}-IV).

The NS-VideoQA framework utilizes SPN to convert an input video into SR. It further utilizes SRM to decompose the compositional question into a program. The program, together with reasoning rules, is iteratively applied to SR, ultimately generating the answer. Further details are described below.
\subsection{Symbolic Representation}
Symbolic representation of static scene. For a video with $T$ frames 
$ v \!=\!\left\{I_1,I_2, \!\ldotp\!\ldotp\!\ldotp\!,I_T\right\}$, 
the Static Scene Parser $\boldsymbol{\varphi}$ parses relationships between the subject-object pairs in video 
$\varphi(v) \!=\!\left\{ e_1, e_2, \!\ldotp\!\ldotp\!\ldotp\!, e_N\right\}$, where $e_i \!=\! \left( s_{sbj},s_{rel},s_{obj}\right),i \!=\!1,\!\ldotp\!\ldotp\!\ldotp\!,N$, 
$s_{sbj}$, $s_{rel}$ and $s_{obj}$ indicate subject, relation and object, respectively. The inference model can effectively filter objects based on the symbolic representation of the relationships in question. The symbolic representation of static scene is defined as $R_{static}(v) \!=\!\varphi(v)$.

Symbolic representation of dynamic scene. For a video with $T$ clips
$ v\!=\!\left\{c_1, c_2, \!\ldotp\!\ldotp\!\ldotp\!, c_T\right\}$, 
the Dynamic Scene Parser $\boldsymbol{\psi}$ generates actions in video $\psi(v)\!=\! \left\{ y_1,\!\ldotp\!\ldotp\!\ldotp\!,y_N\right\}$, where $y_i\!=\!\left( s_{sbj},s_{act},s_{obj},t_{start},t_{end}\right),i\!=\!1,\!\ldotp\!\ldotp\!\ldotp\!,N$,
$s_{sbj}$, $s_{act}$, $s_{obj}$, $t_{start}$ and $t_{end}$ indicate subject, action, object, starting and ending times, respectively. The reasoning model can effectively filter action instances based on the symbolic representation of the actions, it also can accurately determine their temporal order based on the starting and ending times of the actions. The symbolic representation of dynamic scene is defined as $R_{dynamic}(v)\!=\!\psi(v)$. 

Therefore, the SR can be expressed as follows:
\begin{small}
	\begin{equation}
		SR(v)\!=\!\left( R_{static}(v), R_{dynamic}(v)\right) \label{con:eq_1}
	\end{equation}
\end{small}
\subsection{Static-dynamic Scene Parser Network}
\textbf{Static Scene Parser} $\boldsymbol{\varphi}$. Inspired by STTran \cite{cong2021spatial}, the $t$-th frame $I_t$ is fed into Object-Relation Detector \cite{hu2018relation} to detect objects $ o_t^1,\!\ldotp\!\ldotp\!\ldotp\!,o_t^N$, where $o_t^i$ indicates the $i$-th object, $v_t^i$, $c_t^i$ and $r_t^i$ indicate $o_t^i$ corresponding object, semantic and relation feature, respectively.

Relationship representation. It is generated to capture the relationships of subject-object pairs by leveraging object, semantic, and relation features. We choose an object with higher conficence to be ``person", suppose it to be $o_t^1$, its relationship with other objects can be expressed:
\begin{small}
	\begin{equation}
		x_t^i \!=\!\left( W_sv_t^1,W_ov_t^i,W_rf_r,c_t^1,c_t^i,\right), i \!= \!2,\!\ldotp\!\ldotp\!\ldotp\!,N \label{con:eq_2}
	\end{equation}
\end{small}Where $f_r \!=\!\left[r_t^1 \vdots r_t^i\right]$ is the concatenation of relation features of $o_t^1$ and $o_t^i$, $W_s$,$W_o$,$W_r$are the linear matrices for dimension compression. The relationships of the $t$-th frame are represented as $X_t \!=\!\left\{x_t^2, \!\ldotp\!\ldotp\!\ldotp\!, x_t^N\right\}$.

Encoder. In order to encode the spatial relationship representation of subject-object pairs, an $L$-layer Transformer encoder \cite{vaswani2017attention} is utilized. The query $Q$, key $K$, and value $V$ in the encoder share the same input, and the output of the $n$-layer encoder is:
\begin{small}
\begin{equation}
	\begin{gathered}
		X_t^{(n)} \!=\!\operatorname{Att}_{enc}\left(Q \!=\!K \!=\! V \!=\!X_t^{(n-1)}\right) \\
		X_t^{(0)} \!=\!X_t \label{con:eq_3}
	\end{gathered}
\end{equation}
\end{small}

Decoder. A sliding window of size $\eta$ runs over the sequence of spatial relationship representations $\! \left\{X_1^{(L)}, \!\ldotp\!\ldotp\!\ldotp\!, X_T^{(L)}\right\}$, generating $\! Z_j \!=\!\left\{X_j^{(L)}, \!\ldotp\!\ldotp\!\ldotp\!, X_{j+\eta-1}^{(L)}\right\}, j \!\!=\!\!1, \!\ldotp\!\ldotp\!\ldotp\!, T \!-\! \eta \!+\! 1$. The decoder takes $Z_j$ as input, and consists of $L$ stacked identical self-attention layers $\operatorname{Att}_{dec}()$. The decoder can be described as follows:
\begin{small}
	\begin{equation}
		\begin{gathered}
			Z_j^{(1)} \!=\!\operatorname{Att}_{dec}\left(Q \!=\!K \!=\!Z_j+E_{pos}, V \!=\!Z_j\right) \\
			\quad Z_j^{(n)} \!=\!\operatorname{Att}_{dec}\left(Q \!=\!K \!=\!V \!=\!Z_j^{(n-1)}\right) \label{con:eq_4}
		\end{gathered}
	\end{equation}
\end{small}Where $E_{pos}  \!\in\! \mathbb{R}^{\eta \times 1936}$ is a temporal position encoding.

The output of decoder is represented as $Z_j^{(L)}=\left\{z_j^2, \!\ldotp\!\ldotp\!\ldotp\!, z_j^N\right\}$, we predict the relationship $s_{rel}$ between $o_t^1$ and $o_t^i$ from $z_j^i$. Finally, we obtain the symbolic representation of static scene about the given video $v$:
\begin{small}
\begin{equation}
	\begin{gathered}
		R_{static}(v)=\varphi(v)=\bigcup_{t=1}^T \rho_t \\
		\rho_t=\left\{ e_2, \!\ldotp\!\ldotp\!\ldotp\!, e_N\right\} \\
		e_i=\left( s_{sbj}, s_{rel}, s_{obj}\right), i \!=\! 2, \!\ldotp\!\ldotp\!\ldotp\!, N
		 \label{con:eq_5}
	\end{gathered}
\end{equation}
\end{small}

\textbf{Dynamic Scene Parser} $\boldsymbol{\psi}$. Inspired by ActionFormer \cite{zhang2022actionformer}, we utilize the pretrained I3D \cite{carreira2017quo} model to extract clip features $ X \!=\! \left\{x_1, x_2, \!\ldotp\!\ldotp\!\ldotp\!, x_T\right\}$, where $T$ is the number of clips.

Action representation. A projection function embeds each clip feature $x_i$ into a $D$-dimensional space.
\begin{small}
	\begin{equation}
			{E}=\left[\operatorname{Conv}\left(x_1\right), \operatorname{Conv}\left(x_2\right), \!\ldotp\!\ldotp\!\ldotp\!, \operatorname{Conv}\left(x_T\right)\right]^T
			\label{con:eq_6}
	\end{equation}
\end{small}Where $\operatorname{Conv}\left(x_i\right) \in \mathbb{R}^{D}$ consists of $conv1d$ with kernel $size=3$ and $stride=1$ and ReLU. 

Encoder. We utilize encoder ${E}$, to obtain the muti-scale action representation $Z=\left\{Z^{(1)}, \!\ldotp\!\ldotp\!\ldotp\!, Z^{(n)}\right\}$. The query $Q$, key $K$, and value $V$ in the encoder share the same input, and the output of the $n$-layer encoder is:
\begin{small}
\begin{equation}
	\begin{gathered}
		Z^{(n)}=\downarrow\left(\operatorname{Att}_{enc}\left(Q=K=V=Z^{(n-1)}+E_{pos}\right)\right) \\
		Z^{(0)}=E
		\label{con:eq_7}
	\end{gathered}
\end{equation}
\end{small}Where $E_{pos} \!\in\! \mathbb{R}^{D}$ is a position embedding, $\downarrow \!(\! \cdot)$ is the downsampling operator, with a strided depthwise $conv1d$. We use 2x downsampling in this work.

Decoder. The decoder is a lightweight convolutional network with classification and regression heads. It predicts the category, starting and ending time of actions $\left\{ y_1,\!\ldotp\!\ldotp\!\ldotp\!,y_N\right\}$ from $Z$, where $y_i=\left( s_{sbj},s_{act},s_{obj},t_{start},t_{end}\right),i \!=\! 1,\!\ldotp\!\ldotp\!\ldotp\!,N$. The symbolic representation of dynamic scene is represented as:
\begin{small}
	\begin{equation}
		\begin{gathered}
			R_{dynamic}(v)=\psi(v)=\left\{ y_1,\!\ldotp\!\ldotp\!\ldotp\!,y_N\right\}
			\label{con:eq_8}
			\end{gathered}
	\end{equation}
\end{small}

To get the SR of static-dynamic scene, we combine $R_{dynamic}(v)$ and $R_{dynamic}(v)$ according to Eq.\ref{con:eq_1}.

\subsection{Symbolic Reasoning Machine}
SRM decomposes compositional questions with a top-down process and inferences answers with a bottom-up process. To realize the inference of leaf ($ObjectExists$,etc.) and compositional questions ($Choose$, etc.), SRM takes SR and questions as input, and applies symbolic reasoning rules to the SR to obtain the final answer.

\textbf{Language Question Parser}. Same as in \cite{yi2018neural}. To ensure the completeness of NS-VideoQA, we briefly describe the question parser. To obtain program sequence $ p=\left\{p_1, p_2, \!\ldotp\!\ldotp\!\ldotp\!, p_M\right\}$, the question parser utilizes BiLSTM \cite{hochreiter1997long} to encode a question $ q=\left\{q_1, q_2, \!\ldotp\!\ldotp\!\ldotp\!, q_T\right\}$, where $M$ denotes the length of the program, $T$ denotes the length of the question. The middle representation of the encoded question $q$ at time $i$ as:
\begin{small}
\begin{equation}
	\begin{gathered}
		e_i=\left\lceil e_i^F, e_i^B\right\rceil, \text {where }\\ 	e_i^F,h_i^F=\operatorname{LSTM}\left(\Phi_E\left(q_i\right),h_{i-1}^F\right),\\
		e_i^B, h_i^B=\operatorname{LSTM}\left(\Phi_E\left(q_i\right), h_{i+1}^B\right)
		\label{con:eq_9}
	\end{gathered}
\end{equation}
\end{small}Where $\! \Phi_E \!$ is the encoder word embedding. $\left(e_i^F,h_i^F\right)$,$\left(e_i^B,h_i^B\right)$ are the outputs and hidden vectors of the forward and backward networks.

The decoder is a network of LSTM+Attention \cite{bahdanau2014neural}. The decoder generates a token $p_t$ according to the previous token of output sequence $y_{t-1}$, then $p_t$ and $e_i$ are integrated into Attention to obtain a context vector $c_t$, which is a weighted sum of the hidden state:
\begin{small}
	\begin{equation}
		\begin{gathered}
			p_t = \operatorname{LSTM}\left(\Phi_D\left(y_{t-1}\right)\right), 
			\alpha_{ti} \propto \exp \left(p_t^T W_A e_i\right),\\
			c_t =\sum \alpha_{t i} e_i 
			\label{con:eq_10}
		\end{gathered}
	\end{equation}
\end{small}Where $\Phi_D$ is the decoder word embedding. $W_A$ is the attention weight matrix. To obtain the distribution of predicted program 
$p_t \sim \operatorname{softmax}\left(W_O\left\lceil y_t, c_t\right]\right)$, 
the context vector and decoder output are integrated into a fully connected layer with softmax activation.

\textbf{Program Symbolic Executor}. Considering the polymorphism of compositional questions (where the reasoning rules corresponding to the same combination type can vary depending on the category of the sub-question). For a program $p$ with $n$ direct sub-questions 
$C_q=\left\{c_1,\!\ldotp\!\ldotp\!\ldotp\!,c_n\right\}$, assuming $\tau_p$ is the question type of $p$, we define 
$token_p \!:= \! \left\{\tau_p, \tau_{c_1}, \!\ldotp\!\ldotp\!\ldotp\!, \tau_{c_n}\right\}$. Each token corresponds to a reasoning rule $rule_{token}$, with more details in the Supplementary.

The reasoning is conducted by applying  $rule_{token_p}$ iteratively. 
At each step, $rule_{token_p}$ is applied to different inputs determined by the question structure. For leaf questions, the input is $SR(v)$; for compositional questions, the input is the intermediate results from all descendent questions of $q$. 
The result of a question denoted as $a_q \!:= \! \left\{ ans_q, R_q\right\}$, packs two components: a symbolic answer $ans_q \in\left(\{yes,no\} \cup V_{obj} \cup V_{rel} \cup V_{action}\right)$, and symbolic representations $R_q \in SR(v)$ filtered for the question. All intermediate results obtained previously are recorded in a trace $T$, implemented as a dictionary. Thus, the reasoning step for question $q$ becomes:
\begin{small}
\begin{equation}
	a_q= \begin{cases}\operatorname{rule}_{\text {token}_p}(\operatorname{SR}(v)), & q \in Q_l \\ \operatorname{rule}_{\text {token}_p}(T), & q \in Q_c
	\label{con:eq_11}\end{cases}
\end{equation}
\end{small}Where $Q_l$ is leaf question, $Q_c$ is compositional question.

We design a program execution algorithm with dynamic programming, as Algorithm~\ref{alg:execution}. 
For compositional questions, first, the trace is updated iteratively for each sub-question; then, the chosen rule is applied to the trace; finally, the obtained intermediate result is appended to the trace.
\begin{small}
	\begin{algorithm}[H]
		\caption{\bf{Program\_Execution}}
		\label{alg:execution}
		\begin{algorithmic}
		\STATE\textbf{Input:} program $p$, scene representation $SR(v)$, trace $T_{in}$\\
		\STATE\textbf{Output:} trace $T$
		\STATE \# initialization
		\STATE $token \gets \operatorname{get\_token(}p) \quad$  
		\STATE $rule \gets \operatorname{get\_rule}(token)$ 
		\STATE $T \gets T_{in}$
		\STATE \# reasoning  
		\IF{$p$ is leaf question}                                
		\STATE $ a \gets $ rule$(SR(v))$
		\ELSE
		\FOR{$f$ in subprogram of $p$}
		\STATE $T \gets T \cup \operatorname{Program\_Execution}(f,SR(v),T)$
		\ENDFOR
		\STATE $ a \gets $ rule($T$)\\
		\ENDIF
		\STATE \# integration
		\STATE $T \gets T \cup\{$ (token, a $)\} \quad$ 
		\RETURN $T$
		\end{algorithmic}

	\end{algorithm}
\end{small}
\section{Experiments}
\label{sec:Experiments}
To demonstrate the effectiveness of NS-VideoQA, our experimental design is as follows: First, we introduce the dataset and evaluation metrics. Secondly, we analyze the experimental results of the sub-questions, progressively presenting evidence to validate the reasoning capabilities of our model. Finally, we show the visualization process of model reasoning questions.
\begin{figure}[htbp]
	\centering
	\setlength{\belowcaptionskip}{-0.4cm}
	
	\includegraphics[width=85mm]{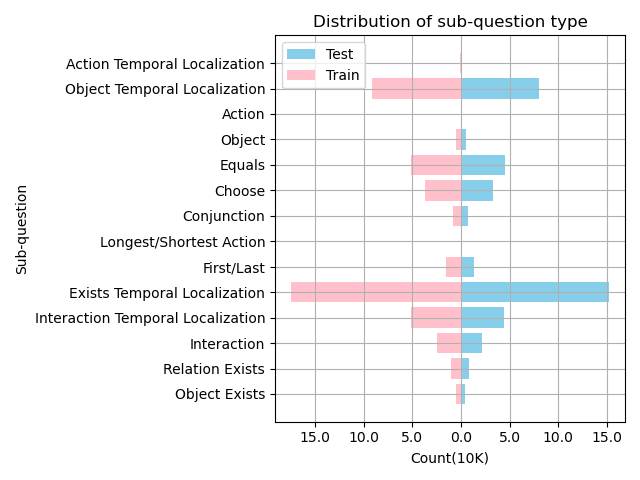}
	\caption{A chart displays the distribution of sub-question types in the train and test sets.
	}
	\label{distribution}
\end{figure}
\subsection{Dataset and Metrics}
\textbf{Dataset.} The dataset of AGQA Decomp\cite{gandhi2022measuring} is organized into 14 distinct sub-question categories, with each category linked to a specific function token. Leveraging indirect references of sub-questions through hand-designed reasoning rules enables the generation of corresponding questions and programs. The specific distribution of sub-question categories for training and testing is shown in Figure~\ref{distribution}.

\textbf{Metrics.} For comprehensive analysis, we study the reasoning capability of our NS-VideoQA via measuring the accuracy, compositional accuracy (CA), right for the Wrong Reasons (RWR), delta, and internal consistency (IC) following AGQA Decomp\cite{gandhi2022measuring}. CA measures the proficiency of the model in compositional reasoning, i.e., to test if the model can answer a question correctly based on correct reasoning results of its sub-questions. RWR scrutinizes the irrationality of reasoning performance by checking if the model guessed the answer correctly but based on wrong intermediate results from sub-questions. Delta represents the discrepancy between RWR and CA, and when RWR surpasses CA with a positive delta value, it indicates incorrect reasoning. IC quantifies the ability to avoid self-contradictions. A higher IC value suggests that the model exhibits compositional consistency satisfying logical reasoning rules. Precise definitions of these indicators and experimental details can be found in the Supplementary.

\subsection{Performance on NS-VideoQA}
Table~\ref{tab1_test_acc} summarizes the comparative results across various sub-question categories. Here, Most-Likely \cite{gandhi2022measuring} generates the most probable answer dependent on language biases. HME \cite{fan2019heterogeneous}, HCRN \cite{le2020hierarchical}, and PSAC \cite{li2019beyond}, are purely neural VideoQA models. Oracle SR represents the reasoning performance of our SRM with ideal SR as inputs. Human is the annotators who agrees with ground-truth answers.

\begin{table}
	\centering
	\caption{We report accuracy values of sub-question type on AGQA Decomp. "-" indicates there were no results for a given type, bold numbers indicate the best accuracy.}
	\label{tab1_test_acc}
	\begin{center}
		\scalebox{0.59}{
			\begin{tabular}{lrrrrrrr}
				\multirow[b]{2}{*}{Sub-question Type} & \multicolumn{7}{c}{Accuracy $\uparrow$} \\
				\cmidrule(r){2-8} 
				& Most-Likely & HME & HCRN & PSAC & Ours & Oracle SR & Human \\
				\hline \hline Object Exists & 50.00 & 45.36 & 31.69 & 88.10 & \textbf{96.26} & 99.92 & 92.00 \\
				Relation Exists & 50.00 & 51.33 & 61.72 & \textbf{100} & 72.94 & 99.00 & 92.00 \\
				Interaction & 50.00 & 50.00 & 50.24 & 41.88 & \textbf{63.35} & 92.24 & 88.00 \\
				\hline Interaction Temporal Loc. & 50.00 & 50.00 & 50.85 & 39.37 & \textbf{50.76} & 73.80 & 96.00 \\
				Exists Temporal Loc. & 50.00 & \textbf{50.00} & 45.27 & 39.55 & - & - & 92.00 \\
				Object Temporal Loc. & - & - & 25.43 & 22.85 & \textbf{35.58}& 68.55 & - \\
				Action Temporal Loc. & - & - & 4.06 & 3.70 & \textbf{67.17} & 92.03 & - \\
				Longest/Shortest Action & 3.57 & 1.79 & 0.00 & 1.79 & \textbf{53.68} & 100 & 76.00 \\
				Action & - & - & 0.00 & 16.67 & \textbf{57.14}& 100 & - \\
				Object & - & - & 26.23 & 18.98 & \textbf{42.64} & 79.69 & - \\
				\hline  Choose & 26.03 & 12.24 & 20.62 & 26.88 & \textbf{31.86}& 84.67 & 88.00 \\
				Equals & 49.92 & & 50.00 & 31.66 & \textbf{53.52} & 63.33& 70.00 \\
				Conjunction & 50.00 & 50.00 & 50.15 & 44.81 & \textbf{51.07}& 53.89 & 76.00 \\
				First/Last & 3.79 & 2.93 & 0.00 & 4.31 & \textbf{13.19}& 46.01 & 88.00 \\
				\hline Overall & 3.31 & 8.28 & 12.43 & 14.98 & \textbf{23.51} & 74.52 & 84.36 \\
			\end{tabular}
		}
	\end{center}
\end{table}

\begin{table*}
	\centering
	\caption{We report compositional accuracy (CA), right for the wrong reasons (RWR), delta (RWR-CA),internal consistency (IC) values. The bold numbers indicate the best results, and the underlined numbers represent the results of the question categories that require attention.N/A indicates there were no valid compositions for a given type.}
	\label{sub_question_tab_2}
	\begin{center}
		\scalebox{0.7}{
			\begin{tabular}{lrrrrrrrrrrrrrrrr}
				\multirow[b]{2}{*}{Sub-question Type} & \multicolumn{4}{c}{$\mathrm{CA}\uparrow$} & \multicolumn{4}{c}{$\mathrm{RWR}\downarrow$} & \multicolumn{4}{c}{$\mathrm {Delta}\downarrow$} & \multicolumn{4}{c}{$\mathrm{IC}\uparrow$} \\
				\cmidrule(r){2-5} \cmidrule(r){6-9}  \cmidrule(r){10-13} \cmidrule(r){14-17}
				& HME & HCRN & PSAC & Ours & HME & HCRN & PSAC & Ours & HME & HCRN & PSAC & Ours & HME & HCRN & PSAC & Ours \\
				\hline \hline Object Exists & N/A& N/A & N/A& N/A & N/A& N/A & N/A& N/A& N/A& N/A & N/A& N/A& N/A& N/A & N/A & N/A \\
				Relation Exists & N/A & 92.86 & \textbf{100} & 85.39 & 0.00 & 52.46 & N/A & N/A & N/A & -40.40 & N/A & N/A & 50.00 & 46.18 & 50.00 & \textbf{50.00} \\
				Interaction & N/A & 67.31 & 71.47 & \textbf{87.13} & \textbf{19.58} & 50.61 & 63.91 & 29.62 & N/A & -16.70 & -7.57 & -\textbf{57.51} & 50.00 & 48.62 & 47.13 & \textbf{98.42} \\
				\hline Interaction Temporal Loc. & \textbf{100} & 53.69 & 53.00 & 38.20 & 50.08 & 49.82 & \textbf{24.22} & 52.20 & -\textbf{49.92}  & -3.87 & -28.78 & 14.00 & 31.25 & 32.37 & 27.09 & \textbf{32.40} \\
				Exists Temporal Loc. & 0.00 & \textbf{94.79} & 89.38 & - & \textbf{1.26} & 45.45 & 30.33 & - & 1.26 & -49.34 & -\textbf{59.05} & - & \textbf{98.28} & 33.31 & 27.10 & - \\
				Object Temporal Loc. & 60.00 & 37.55 & 35.10 & \textbf{80.46} & 24.75 & 29.36 & 23.91 & \textbf{15.08} & -35.25 & -8.19 & -11.20 & -\textbf{63.58} & \textbf{77.04} & 56.71 & 39.70 & 33.65 \\
				Action Temporal Loc. & N/A & \textbf{8.74} & 8.46 & N/A & \textbf{1.62} & 7.88 & 2.50 & 66.85 & N/A & -0.86 & -\textbf{5.96} & N/A & N/A& 52.65 & 15.16 & \textbf{66.10} \\
				Longest/Shortest Action &N/A & N/A & N/A & N/A & N/A& N/A & N/A & N/A & N/A & N/A & N/A & N/A & N/A & N/A & N/A & N/A\\
				Action & N/A & 0.00 & 28.57 & \textbf{57.14} & 0.00 & 0.00 & N/A & N/A & N/A & 0.00 & N/A & N/A& 14.29 & 14.92 & 14.29 & \textbf{15.38} \\
				Object & N/A& 30.09 & 28.34 & \textbf{46.22} & 24.07 & 34.33 & N/A & \textbf{0.02} & N/A & 4.23 &N/A & -\textbf{46.20} & 14.29 & 14.67 & 14.29 & \textbf{24.24} \\
				\hline Choose & N/A& 39.12 & \textbf{42.96} & $\underline{39.07}$ & 43.28 & 44.75 & 48.57 & $\underline{\textbf{31.26}}$ & N/A & 5.62 & 5.61 & -$\underline{\textbf{7.81}}$ & 0.00 & 13.7 & 16.73 & $\underline{\textbf{49.49}}$ \\
				Equals & 46.47 & 30.94 & 31.65 & $\underline{\textbf{91.67}}$ & 50.58 & 50.63 & \textbf{31.25} & $\underline{49.52}$ & 4.11 & 19.69 & -0.40 &  -$\underline{\textbf{42.14}}$ & 48.9 & 20.77 & 1.36 & $\underline{\textbf{46.93}}$ \\
				Conjunction & 70.00 & 70.23 & \textbf{73.67}& $\underline{63.33}$ & 37.53 & 41.70 & 35.51 & $\underline{\textbf{23.72}}$ & -32.47 & -28.53 & -38.17 &  -$\underline{\textbf{39.61}}$ & \textbf{50.00} & 38.71 & 19.56 & $\underline{48.52}$ \\
				First/Last & 47.01 & 0.03 & 70.66 & $\underline{\textbf{81.66}}$ & 10.02 & \textbf{0.01} & 7.79 & $\underline{3.64}$ & -36.99 & -0.02 & -62.87 &  -$\underline{\textbf{78.02}}$ & 36.80 & 0.02 & \textbf{75.56} & $\underline{26.65}$ \\
				\hline Overall & 23.10 & 37.52 & 45.23 & \textbf{51.44} & \textbf{18.77} & 29.07 & 19.14 & 23.00 & -4.33 & -8.45 & -26.09 & -\textbf{28.44} & 35.67 & 26.62 & 24.85 & \textbf{37.04} \\
			\end{tabular}
		}
	\end{center}
\end{table*}

\begin{table*}
	\centering
	\caption{We report the performances of HME, HCRN, and PSAC on the RWR-$n$ metrics (the lower the better), where $n$ represents the number of incorrectly answered sub-questions for a composition, conditioned on the given parent question type. N/A occurs when none sub-question of this type has $n$ sub-questions misanswered by that model.}
	\label{sub_question_tab_3}
	\begin{center}
		\scalebox{0.515}{
			\begin{tabular}{lrrrrrrrrrrrrrrrrrrrr}
				\multirow{2}{*}{ Sub-question Type } & \multicolumn{5}{c}{ HME } & \multicolumn{5}{c}{ HCRN } & \multicolumn{5}{c}{ PSAC } & \multicolumn{5}{c}{ Ours } \\
				\cmidrule(r){2-6} \cmidrule(r){7-11}  \cmidrule(r){12-16} \cmidrule(r){17-21}
				& RWR-1 & RWR-2 & RWR-3 & RWR-4 & RWR-5 & RWR-1 & RWR-2 & RWR-3 & RWR-4 & RWR-5 & RWR-1 & RWR-2 & RWR-3 & RWR-4 & RWR-5 & RWR-1 & RWR-2 & RWR-3 & RWR-4 & RWR-5 \\
				\hline \hline  Object Exists & N/A& N/A & N/A & N/A & N/A & N/A & N/A& N/A & N/A& N/A& N/A& N/A& N/A & N/A& N/A& N/A& N/A & N/A & N/A & N/A\\
				Relation Exists & 0.00 & N/A & N/A & N/A & N/A & 52.46 & N/A & N/A & N/A& N/A& N/A& N/A& N/A & N/A& N/A & N/A & N/A & N/A & N/A & N/A\\
				Interaction & 28.30 & 46.88 & 11.76 & N/A & N/A & 62.29 & 52.78 & 27.33 & N/A  & N/A& 63.91 & N/A  & N/A & N/A & N/A & 31.71 & 19.68 & N/A & N/A& N/A\\
				\hline Interaction Temporal Loc. & 100 & 99.94 & 70.26 & 99.63 & 3.48 & 45.36 & 56.64 & 60.68 & 49.10 & 6.88 & 26.12 & 21.27 & 24.59 & 24.27 & 11.31 & $\underline{36.59}$ &$\underline{27.71}$& $\underline{70.03}$ & $\underline{69.78}$ & $\underline{10.04}$ \\
				Exists Temporal Loc. & 4.22 & 1.19 & N/A & N/A & N/A & 65.05 & 17.55 & N/A & N/A& N/A& 32.02 & 9.20 & N/A & N/A& N/A& 0.06 & 18.54 & N/A & N/A & N/A\\
				Object Temporal Loc. & 45.86 & 24.80 & 18.55 & 2.63 & N/A & 35.84 & 27.39 & 10.74 & 4.74 & N/A  & 35.30 & 14.49 & 0.00 & N/A  & N/A& - & - & - & - & -  \\
				Action Temporal Loc. & 1.62 & N/A & N/A & N/A & N/A & 7.88 & N/A & N/A& N/A& N/A& 2.50 & N/A  & N/A & N/A& N/A& 66.85 & N/A & N/A& N/A & N/A\\
				Longest/Shortest Action & N/A& N/A& N/A& N/A& N/A& N/A& N/A & N/A & N/A& N/A& N/A & N/A & N/A & N/A & N/A & N/A & N/A & N/A & N/A & N/A\\
				Action & 0.00 & N/A &N/A & N/A & N/A & 0.00 & N/A & N/A& N/A& N/A& N/A& N/A& N/A & N/A& N/A& N/A & N/A & N/A & N/A & N/A\\
				Object & 36.13 & 23.92 & N/A & N/A & N/A & 30.08 & 38.09 & N/A & N/A& N/A & N/A& N/A& N/A & N/A& N/A& 0.02 & N/A & N/A & N/A & N/A\\
				\hline Choose & 44.35 & 18.94 & N/A &N/A & N/A & 44.80 & 44.03 & N/A & N/A & N/A & 55.60 & 32.46 & N/A & N/A & N/A & 29.26 & 15.62 & N/A & N/A & N/A\\
				Equals & 58.21 & 49.10 & N/A& N/A & N/A & 50.64 & 50.62 & N/A & N/A& N/A & 36.01 & 20.78 & N/A & N/A & N/A & 50.54 & 45.29 & N/A & N/A & N/A\\
				Conjunction & 46.81 & 24.35 & N/A & N/A & N/A & 49.27 & 24.74 & N/A & N/A & N/A & 51.58 & 6.35 & N/A & N/A & N/A& 44.89 & 79.46 & N/A  & N/A& N/A\\
				First/Last & 10.02 & N/A & N/A & N/A & N/A & 0.01 & N/A  & N/A& N/A& N/A& 7.79 & N/A & N/A& N/A& N/A& 3.64 & N/A  & N/A & N/A& N/A\\					
				\hline Overall & 38.97 & 19.14 & 40.78 & 54.57 & 38.97 & 3.48 & 35.99 & 45.88 & 38.43 & 6.88 & 38.01 & 20.85 & 24.55 & 24.27 & 11.31 & 22.54 & 22.02 & 54.56 & 49.54 & 10.04 \\
			\end{tabular}
		}
	\end{center}
\end{table*}
Our model achieves an average accuracy of $77.63\%$ on leaf questions ($Object \  Exists$ and $Relation \  Exists$, and $Interaction$ categories). HCRN and HME achieve approximately $50\%$ accuracy, while PSAC achieves an average accuracy of $76.66\%$. We conjecture that the accuracy disparity between purely neural models and neural-symbolic models remains minor in the absence of complex, multi-hop questions. To further investigate this idea, we provide accuracy results for each binary question type in the Supplementary. For the $Temporal\  Loc.$ category, NS-VideoQA exhibits higher accuracy compared to HCRN, HME, and PSAC, especially in $Action\  Temporal\  Loc.$, $Longest/Shortest\  Action$, and $Action$ categories. This can be attributed to the limited action detection and temporal localization capabilities of the above models.

For more complex categories, such as $Choose$, $Equals$, $Conjunction$, and $First/Last$ categories, NS-VideoQA demonstrates superior performance compared to HCRN, HME, and PSAC by approximately $1\%$ to $7\%$. The $Choose$ type requires selecting objects or actions, while the $Equals$ type involves evaluating the similarity between two objects or actions. The $Conjunction$ type consists of And and Xor operations, and the $First/Last$ type involves identifying the first/last object or action. Across these four types, NS-VideoQA consistently outperforms the other models, highlighting its strong capabilities in compositional spatio-temporal reasoning.

In the $Overall$ results, NS-VideoQA achieves an accuracy of $23.51\%$, which is still far from human-level performance. However, it significantly improves the ability of compositional spatio-temporal reasoning compared to other models. To further analyze the reasoning capability of NS-VideoQA, we consider additional metrics, namely CA, RWR, and IC, as shown in Table~\ref{sub_question_tab_2}. These metrics provide a more comprehensive understanding of NS-VideoQA's performance and its effectiveness in capturing and reasoning about complex spatio-temporal information.

\subsection{Performance on Choose, Equals, Conjunction, and First/Last}
To further investigate the reasoning challenges of NS-VideoQA, we assess its ability to achieve high accuracy, and the probability of guessing the answer correctly based on incorrect reasoning. This evaluation is conducted using the CA, RWR, and Delta metrics, as presented in Table~\ref{sub_question_tab_2}.

In the $Choose$ type, NS-VideoQA exhibits a CA value of $39.07\%$, indicating that the model struggles with this category. Although its RWR value of $31.26\%$ does not surpass HCRN and PSAC models, the Delta value of -$7.81\%$ suggests that NS-VideoQA is more likely to perform correct reasoning. Table~\ref{sub_question_tab_3} further validates this point, as NS-VideoQA exhibits lower RWR-1 and RWR-2 values compared to HCRN and PSAC. On the other hand, HCRN and PSAC have CA values higher than NS-VideoQA by $0.05\%$ and $3.89\%$, respectively. However, their positive Delta values of $5.62\%$ and $5.61\%$ indicate that these models tend to guess answers correctly based on incorrect reasoning. The values of CA, RWR, and Delta reveal the challenges faced by HME, HCRN, and PSAC in $Equals$ type, with values of ($46.47\%$, $30.94\%$, $31.56\%$), ($50.58\%$, $50.63\%$, $31.25\%$), and ($4.11\%$, $19.69\%$, -$0.4\%$) respectively. In contrast, our model achieves $91.67\%$, $49.52\%$, and -$42.14\%$ respectively, indicating that NS-VideoQA has compositional spatio-temporal reasoning capability. Similar conclusions can be drawn for the $First/Last$ type.

\begin{figure}[htbp]
	\centering
	\vspace{-0.3cm}
	\setlength{\belowcaptionskip}{-0.3cm}
	\includegraphics[width=85mm]{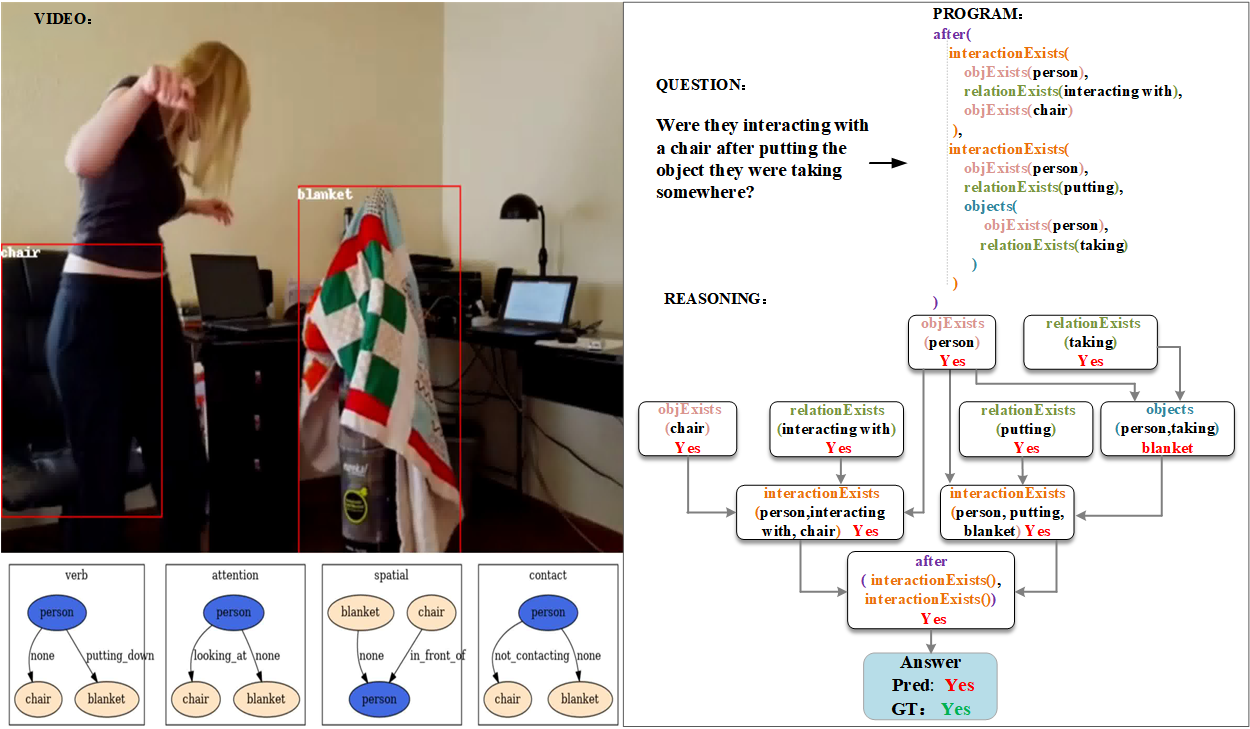}
	
	\caption{The reasoning for the $ Interaction\ Temporal\ Loc.$ type. \textbf{Top left:} the red boxes denote the detected objects ``chair" and ``blanket". \textbf{Bottom left:} the static symbolic representation of verb, attention, spatial, and contact. \textbf{Right:} the answer ``Yes" is obtained by reasoning based on the program.
	}
	\label{example1}
	
\end{figure}
HME, HCRN, and PSAC demonstrate Delta values of -$32.47\%$, -$28.53\%$, and -$38.17\%$, respectively, indicating their proficiency in correctly reasoning on $Conjunction$ compared to $Choose$ and $Equals$. NS-VideoQA exhibits a Delta value of -$39.61\%$, suggesting that errors in intermediate reasoning have a minimal negative impact on its overall performance. This finding further highlights the robustness of NS-VideoQA based on reasoning rules and avoid relying on data biases. Additionally, we assess the model's self-consistency by evaluating the IC metrics in Table~\ref{sub_question_tab_2}. To further prove the effectiveness of NS-VideoQA, we report the results of compositional types in Supplementary.

\subsection{Visualization of execution traces and static-dynamic SR}
\begin{figure}[htbp]
	\centering
	\vspace{-0.3cm}
	\setlength{\belowcaptionskip}{-0.4cm}
	\includegraphics[width=85mm]{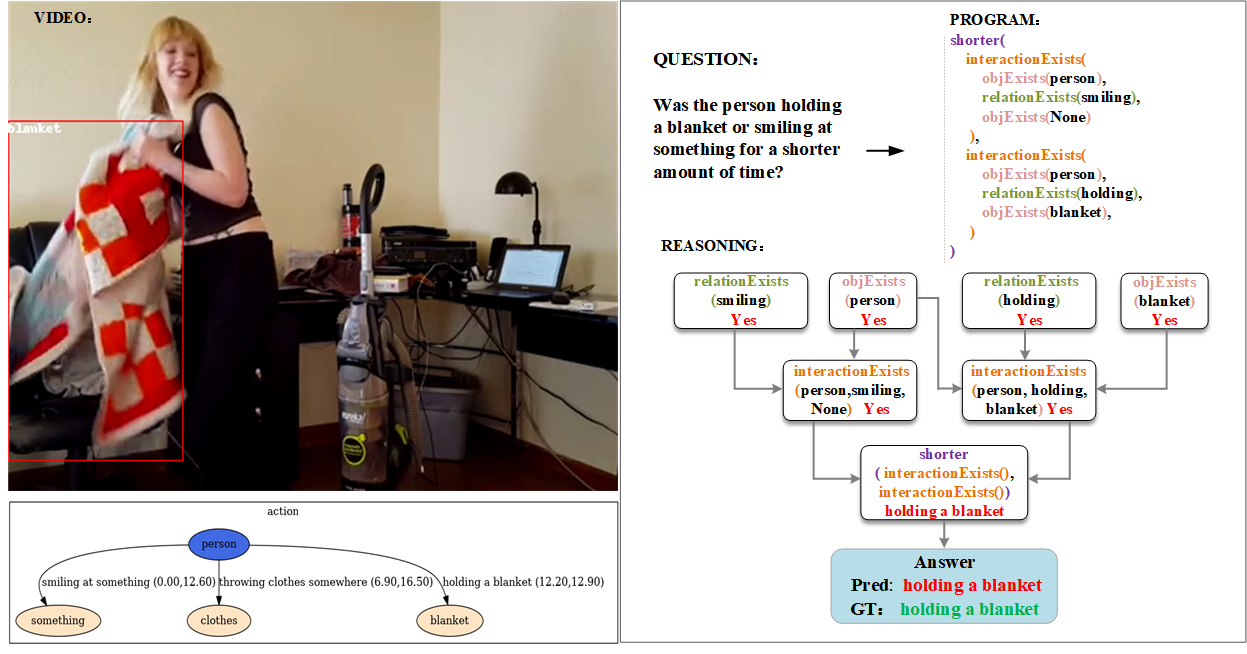}
	
	\caption{The reasoning for the $ Shortest\ Action $ type. \textbf{Top left:} the red box denotes the detected object ``blanket". \textbf{Bottom left:} the symbolic representation of dynamic scene. \textbf{Right:} the answer ``hold a blanket" is obtained by reasoning based on the program.
	}
	\label{example2}
	
\end{figure}

The inherent visual interpretability of NS-VideoQA provides insights into SPN learning and SRM reasoning, leading to a detailed understanding of the mechanism inside. To illustrate, we visualize two instances of inference process as Figure~\ref{example1} and Figure~\ref{example2}. The $ Interaction\  Temporal\  Loc.$ type shown in Figure~\ref{example1} shows the reasoning involved with static SR. In the question, the phrase ``they were taking" become a constraint to ``the object". With the relationship (person, taking, blanket) identified by SPN in previous frames, the SRM clarifies that ``the object" refers to ``blanket". Thus, in the next reasoning step, the action description ``putting the object" instantized to ``putting the blanket", enables the SRM to query the correct action instance from SR for the subsequent temporal reasoning. 

Figure~\ref{example2} shows the temporal reasoning appertain to dynamic SR. For two action instances specified by the given question, the dynamic SR represents their temporal groundings. By calculating and comparing their length, the SRM chooses the shorter one as the final answer in response to $Action\ Shorter $ type. 
The above examples demonstrate: (i) The reasoning rules are conditioned by sub-question types and intermediate results. Thus, the polymorphism of the program executor is necessary. (ii) The reasoning process depends on both static and dynamic scene representations. Without any of them, the answer could not be deduced correctly. (iii)  From the reasoning process, NS-VideoQA enjoys compositional spatio-temporal reasoning ability, interpretability, and generates human-readable execution traces.

\section{Conclusion}
In this paper, the proposed NS-VideoQA improves the capability of compositional spatio-temporal reasoning in real-world by transforming the video into symbolic representation and conducting iterative reasoning. The proposed model performed extensive experiments on AGQA Decomp dataset, the CA, RWR, Delta, and CA results show that NS-VideoQA exhibits internal consistency and significantly improves the accuracy of compositional spatio-temporal questions. The quantitative and qualitative analysis leads to the following conclusions:
(i) In simple questions without complex spatio-temporal reasoning, neural-symbolic and purely neural methods show minimal differences. However, in complex scenarios requiring compositional spatio-temporal reasoning, purely neural VideoQA methods often rely on data biases for answer generation, lacking robust reasoning capabilities.
(ii) Accurate static symbolic representations clarify relationships and filter objects, while accurate dynamic symbolic representations indicate action chronologies, aiding precise localization of behaviors and action sequencing.

The NS-VideoQA has certain limitations. When the symbolic representations are inaccurate (\eg, low accuracy in object detection, relationship recognition, and action localization), the accuracy of question answering in NS-VideoQA can be affected. Additionally, even if the symbolic representations are accurate, the reasoning capabilities of NS-VideoQA can be hindered if the reasoning rules are ambiguous or prone to ambiguity. We hope our paper could motivate future research in obtaining accruate symbolic representations and generating reasoning rules with more unambiguity.

\clearpage
{
    \small
    \bibliographystyle{unsrt}
    \bibliography{main.bib}
}
\clearpage
\appendix
\section*{Supplementary}
The supplementary sections provide more detail on the methods and experiments described in our paper.

To help results reproduction, we report:
\begin{itemize}[leftmargin=0.85cm]
	\item Implementation details of NS-VideoQA in~\ref{sec:Implementation},
	\item Feature extraction process in ~\ref{sec:Feature},
	\item Training methods in ~\ref{sec:Training},
	\item Details of metrics in ~\ref{sec:Metrics}.
\end{itemize}

Addition to the experimental results, we report:
\begin{itemize}[leftmargin=0.85cm]
	\item Performance on composition questions in ~\ref{sec:Performance},
	\item Extra examples in ~\ref{sec:Examples},
	\item Reasoning rules of AGQA Decomp in ~\ref{sec:Reasoning}.
\end{itemize}

\section{Implementation Details}
\label{sec:Implementation}
We report our implementation details including training settings, hyper-parameters, and input/output dimensions for each part of our NS-VideoQA. The implementation details for Object-Relation Detector, Faster RCNN, ActionFormer and Bi-LSTM Encoder-Decoder are listed in Table~\ref{tab4_parames_static}, Table~\ref{tab5_parames_faster}, Table~\ref{tab6_parames_dynamic}, and Table~\ref{tab7_parames_SRM} respectively.

\section{Feature Extraction}
\label{sec:Feature}
\textbf{Static Scene Parser.} Faster R-CNN based on ResNet101 is used as object detection backbone, and Relation Network \cite{hu2018relation} is applied to the instance recognition stage of Faster R-CNN. Each frame $I_t$ is fed into Faster R-CNN to detect 100 objects, and obtain the 2048D object feature vectors, bounding boxes, and class distributions. The class distributions are linearly transformed into 200D semantic features by a linear matrix  $W_s  \!\in\! \mathbb{R}^{36 \times 200}$. The instance recognition stage computes relation features based on each object feature and bounding box. 

For the 100 objects detected on the $t$-th frame $\left\{v_t^i, b_t^i\right\}_{i=1}^{100}$, the object weight $w_{t,v}^{ji}$ and box weight $w_{t,b}^{ji}$ of the $i$-th object with respect to any other object $j$ are calculated as follows:

\begin{small}
	\begin{equation}
		\begin{gathered}
			w_{t, v}^{j i}=\frac{\operatorname{dot}\left(W_v v_t^j, W_q v_t^i\right)}{\sqrt{d_k}} \label{con:eq_12}
		\end{gathered}
	\end{equation}
\end{small}
\begin{small}
	\begin{equation}
		\begin{gathered}
		w_{t, b}^{j i}=\max \left\{0, W_b \cdot \varepsilon_b\left(b_t^j, b_t^i\right)\right\} \label{con:eq_13}
		\end{gathered}
	\end{equation}
\end{small}
\begin{small}
	\begin{equation}
		\begin{gathered}
			\varepsilon_b\!\left(\!b_t^j, b_t^i\!\right)\!\!=\!\!\left[\!\log \!\left(\!
			\frac{\left|x_j-x_i\right|}{w_j}, 
			\frac{\left|y_j-y_i\right|}{h_j},
			\frac{w_i}{w_j}, 
			\frac{h_i}{h_j}\!\right)\!\!\right]^T\! \label{con:eq_14}
		\end{gathered}
	\end{equation}
\end{small}Where $\sqrt{d_k}$ is scaling factor. Finally, the relation feature $r_t^i$ is computed as:

\begin{table}
	\centering
	\caption{Training settings, hyper-parameters, and input/output dimension of the Object-Relation Detector in Static Scene Parser.}
	\label{tab4_parames_static}
	\begin{center}
		\scalebox{0.75}{
			\begin{tabular}{c|cc}
				\hline
				Object-Relation Detector & key & value \\
				\hline
				\multirow{6}{*}{ training settings } & learning rate & $10^{-4}$ \\
				& optimizer & AdamW \\
				& epoch & 10 \\
				& batch size & 10 frames\\
				& weight decay & $10^{-3}$ \\
				& dropout & 0.1 \\
				\hline
				\multirow{4}{*}{hyper-parameters} & encoder layer & 1 \\
				& decoder layer & 3 \\
				& window size $\eta$ for transformer & 2 \\
				& window stride for transformer & 1 \\
				\hline
				\multirow{4}{*}{ input/output dimension } & object feature length & 2048D \\
				& relation feature length & 2048D \\
				& semantic feature length & 200D \\
				& feature length for transformer & 1936D \\
				\hline
			\end{tabular}
		}
	\end{center}
\end{table}

\begin{table}
	\centering
	\caption{Training settings, hyper-parameters, and input/output dimension of the Faster RCNN in Static Scene Parser.}
	\label{tab5_parames_faster}
	\begin{center}
		\scalebox{0.89}{
			\begin{tabular}{c|cc}
				\hline
				Faster RCNN & key & value \\
				\hline
				\multirow{5}{*}{ training settings } & learning rate & $10^{-4}$ \\
				& optimizer & SGD \\
				& epoch & 20 \\
				& batch size & 4 frames\\
				& weight decay & $10^{-4}$ \\
				\hline
				{hyper-parameters} & object numbers & 100 \\
				\hline
				\multirow{3}{*}{ input/output dimension } & feature map length & 1024D \\
				& object feature length & 2048D \\
				& relation feature length & 2048D \\
				\hline
			\end{tabular}
		}
	\end{center}
\end{table}

\begin{table}
	\centering
	\caption{Training settings, hyper-parameters, and input/output dimension of the ActionFormer in Dynamic Scene Parser.}
	\label{tab6_parames_dynamic}
	\begin{center}
		\scalebox{0.92}{
			\begin{tabular}{c|cc}
				\hline ActionFormer & key & value \\
				\hline
				\multirow{8}{*}{ training settings } & learning rate & $10^{-4}$ \\
				& optimizer & Adam \\
				& epoch & 50 \\
				& warmup epoch & 5 \\
				& batch size & 2 clips \\
				& weight decay & 0.05 \\
				& loss coefficient $\lambda$ & 1 \\
				& dropout & 0.1 \\
				\hline
				\multirow{2}{*}{ hyper-parameters } & feature pyramid layer & 7 \\
				& downsampling ratio & 2 \\
				\hline
				input/output dimension & motion feature length & 1024D \\
				\hline
			\end{tabular}
		}
	\end{center}
\end{table}

\begin{table}
	\centering
	\caption{Training settings, hyper-parameters, and input/output dimension of the Bi-LSTM Encoder-Decoder in Symbolic Reasoning Machine.}
	\label{tab7_parames_SRM}
	\begin{center}
		\scalebox{0.77}{
			\begin{tabular}{c|cc}
				\hline
				Bi-LSTM Encoder-Decoder & key & value \\
				\hline
				\multirow{5}{*}{ training settings } & learning rate & $7 \times 10^{-4}$ \\
				& optimizer & Adam \\
				& iteration & 21721 \\
				& batch size & 64 questions\\
				& weight decay & 0.9 \\
				\hline
				\multirow{5}{*}{hyper-parameters} & hidden layer & 2 \\
				& hidden layer dimension& 256D \\
				& encoder layer & 1 \\
				& decoder layer & 1 \\
				& GLoVe word embedding & 300D \\
				\hline
				\multirow{2}{*}{ input/output dimension } & length of question sequence & 49D \\
				& length of program sequence & 81D \\
				\hline
			\end{tabular}
		}
	\end{center}
\end{table}

\begin{small}
	\begin{equation}
		\begin{gathered}
			w_t^{j i}=\frac{w_{t, b}^{j i} \cdot \exp \left(w_{t, v}^{j i}\right)}{\sum\nolimits_k w_{t, b}^{k i} \cdot \exp \left(w_{t, v}^{k i}\right)} \label{con:eq_15}
		\end{gathered}
	\end{equation}
\end{small}
\begin{small}
	\begin{equation}
		\begin{gathered}
			r_t^i \!=\! \sum\nolimits_{j} w_t^{j i} \cdot\left(W_v v_t^j\right) \label{con:eq_16}
		\end{gathered}
	\end{equation}
\end{small}

The parameters of static scene parser including RPN are fixed when training scene graph generation models.

\textbf{Dynamic Scene Parser.} The motion features are obtained by passing RGB clips through an I3D \cite{carreira2017quo} network pre-trained on Kinetics400. Following the experimental setup in ActionFormer \cite{zhang2022actionformer}, we first resize the frames to scale the shortest side to 256; then crop to $224 \times 224$ at center. The clips are obtained by a sliding window with size 16 and stride 4. For each clip, we extract the activation value of the group 5 in the I3D network and apply a global average pooling to obtain a feature vector of 1024D.

\section{Training NS-VideoQA}
\label{sec:Training}
\textbf{Training SPN.} The Static Scene Parser and the Dynamic Scene Parser in the SPN can be trained separately, such modular architecture makes SPN compatible to the improvements of relation recognition and temporal action detection methods in future.

In the Static Scene Parser, a cross-entropy loss $L_{obj}$ is applied to train object detection, and a multi-label margin loss $L_r$ is introduced to optimize the detection of multiple relation for the same subject-object pair. For a subject-object pair $p$, let $R^+$ and $R^-$ be the relation set exists and not exists between $p$ in ground-truth, let $\alpha(p, r)$ be the predicted probability that the subject-object pair $p$ has relation $r$, then the $L_r$ is calculated as:
\begin{small}
	\begin{equation}
		\begin{gathered}
			L_r\left(p, R^{+}, R^{-}\right) \!=\sum_{r \in R^{+}} \sum_{r^{\prime} \in R^{-}}  \!\!\max\! \left(0,1-\alpha(p, r)+\alpha\left(p, r^{\prime}\right)\right) \label{con:eq_17}
		\end{gathered}
	\end{equation}
\end{small}

The total loss $L_s$ of the Static Scene Parser is:
\begin{small}
	\begin{equation}
		\begin{gathered}
			L_s = L_r +L_{obj}  \label{con:eq_18}
		\end{gathered}
	\end{equation}
\end{small}

In the Dynamic Scene Parser, for each action, its category and start/end time are predicted by classification and regression, respectively. Thus, the classification loss $L_{cls}$ for the action category is a Focal loss, and the regression loss $L_{reg}$ for the action start/end time is a DIoU loss.

The total loss $L_d$ of the Dynamic Scene Parser is:
\begin{small}
	\begin{equation}
		\begin{gathered}
			L_d=\sum_L \sum_t\left(L_{c l s}+\lambda(0,1) L_{\text {reg }}\right) / T^{+}  \label{con:eq_19}
		\end{gathered}
	\end{equation}
\end{small}Where $L$ refers to the number of layers in the feature pyramid, $T^+$ is the number of positive samples, $\lambda \!=\!1$ is a coefficient to balance the classification and regression.

\textbf{Training SRM.} In SRM, a Bi-LSTM Encoder-Decoder is trained as the Language Question Parser to generate the programs from questions, where each element of the output sequence refers to a program statement corresponding to a reasoning step. We divide all program statements into a total of $J\!=\!1210$ categories according to their question type and sub-question types. Treating the prediction of each element as a classification problem, the loss function of the Language Question Parser is an element-by-element cross-entropy loss:
\begin{small}
	\begin{equation}
		\begin{gathered}
			L_{prog }=-\sum_{j=1}^J \mathbb{I}\left\{y_p=j\right\} \log \left(p_j\right)  \label{con:eq_20}
		\end{gathered}
	\end{equation}
\end{small}Where, $p_j$ is the predicted probability that the current element belongs to the $j$-th category, $y_p$ is the ground-truth label of the current element.

\begin{figure}[htbp]
	\centering
	\setlength{\belowcaptionskip}{-0.4cm}
	
	\includegraphics[width=85mm]{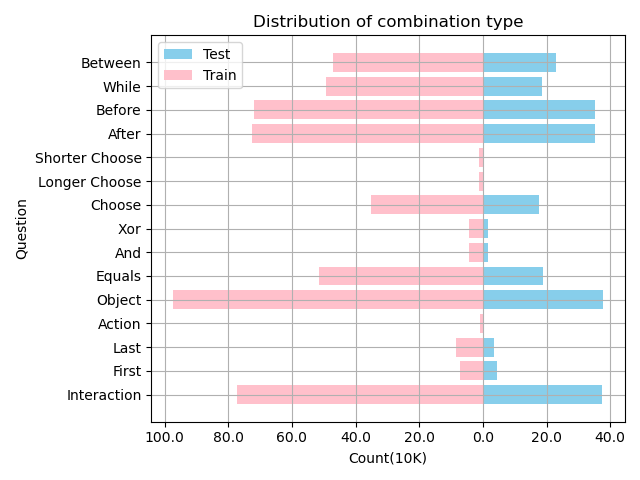}
	\caption{A chart displays the distribution of combination types in the train and test sets.
	}
	\label{distribution_combination}
\end{figure}

\begin{table*}
	\centering
	\caption{We report compositional accuracy (CA), right for the wrong reasons (RWR), delta (RWR-CA), internal consistency (IC) values respect to composition rules for HME, HCRN, PSAC and Ours model.}
	\label{tab_8_combination_CA}
	\begin{center}
		\scalebox{0.73}{
			\begin{tabular}{lrrrrrrrrrrrrrrrr}
				\multirow[b]{2}{*}{Composition Type} & \multicolumn{4}{c}{$\mathrm{CA}\uparrow$} & \multicolumn{4}{c}{$\mathrm{RWR}\downarrow$} & \multicolumn{4}{c}{$\mathrm {Delta}\downarrow$} & \multicolumn{4}{c}{$\mathrm{IC}\uparrow$} \\
				\cmidrule(r){2-5} \cmidrule(r){6-9}  \cmidrule(r){10-13} \cmidrule(r){14-17}
				& HME & HCRN & PSAC & Ours & HME & HCRN & PSAC & Ours & HME & HCRN & PSAC & Ours & HME & HCRN & PSAC & Ours \\
				\hline \hline
				Interaction & 100 & 56.12 & 56.07 & 77.43 & 36.71 & 52.65 & 41.75 & 50.75 & -63.29 & -3.47 & -14.31 & -26.68 & 50.00 & 48.61 & 39.69 & 61.14 \\
				First & 48.16 & 0.02 & 72.66 & 92.63 & 8.62 & 0.01 & 7.90 & 1.72 & -39.54 & -0.01 & -64.77 & -90.91 & 37.90 & 0.03 & 74.96 & 52.12 \\
				Last & 45.61 & 0.04 & 68.19 & 70.89 & 11.93 & 0.01 & 7.65 & 6.36 & -33.68 & -0.04 & -60.55 & -64.54 & 35.70 & 0.02 & 76.15 & 22.47 \\
				 Action & N/A & 8.61 & 8.88 & 57.14 & 1.59 & 7.69 & 2.50 & 26.95 & N/A & -0.92 & -6.38 & -30.19 & 42.37 & 34.12 & 25.79 & 42.08 \\
				Object & N/A & 33.63 & 31.90 & 31.64 & 24.73 & 32.47 & 27.56 & 15.04 & N/A & -1.16 & -4.34 & -16.6 & 58.31 & 48.73 & 36.97 & 50.40 \\
				Equals & 46.47 & 30.94 & 31.65 & 91.67 & 50.58 & 50.63 & 31.25 & 49.52 & 4.11 & 19.69 & -0.40 & -42.14 & 48.90 & 20.77 & 1.36 & 46.93 \\
				And & 100 & 89.16 & 96.81 & 54.42 & 41.28 & 27.67 & 14.08 & 48.77 & -58.72 & -61.49 & -82.73 & -5.65 & 50.00 & 61.78 & 39.13 & 51.02 \\
				Xor & 100 & 38.40 & 0.00 & 94.81 & 33.26 & 53.04 & 50.79 & 40.60 & -66.74 & 14.63 & 50.79 & -54.21 & 50.00 & 15.64 & 0.00 & 82.18 \\
				Choose & N/A & 40.61 & 80.95 & 86.85 & 44.36 & 45.08 & 48.57 & 31.26 & N/A & 4.47 & -32.38 & -55.59 & 0.00 & 13.70 & 16.73 & 49.19 \\
				Longer Choose & N/A & 33.68 & 42.65 & 0.00 & 19.19 & 33.78 & N/A & N/A & N/A & 0.10 & N/A & N/A & N/A & N/A & N/A & N/A \\
				Shorter Choose & N/A & 33.96 & 43.08 & 0.00 & 18.26 & 32.34 & N/A & N/A & N/A & -1.62 & N/A & N/A & N/A & N/A & N/A & N/A \\
				After & 72.73 & 76.59 & 72.05 & 23.80 & 21.86 & 49.53 & 23.89 & 21.87 & -50.87 & -27.06 & -48.15 & -1.93 & 68.87 & 52.11 & 38.46 & 38.67 \\
				Before & 63.16 & 77.50 & 71.70 & 21.69 & 21.27 & 47.66 & 23.69 & 20.49 & -41.89 & -29.84 & -48.01 & -1.20 & 68.84 & 51.89 & 35.76 & 48.30 \\
				While & 75.00 & 89.99 & 83.03 & 7.80 & 8.43 & 47.20 & 26.42 & 7.49 & -66.57 & -42.79 & -56.61 & -0.31 & 59.45 & 54.9 & 46.24 & 27.92 \\
				Between & 95.83 & 92.91 & 98.18 & 94.28 & 6.60 & 18.73 & 16.93 & 15.95 & -89.23 & -74.18 & -81.25 & -78.32 & 58.50 & 68.20 & 42.57 & 29.16 \\
				\hline Overall & 86.52 & 62.25 & 62.63 & 36.40 & 26.75 & 43.13 & 32.15 & 27.41 & -59.76 & -19.13 & -30.48 & -8.09 & 48.92 & 40.34 & 34.84 & 42.53 \\
			\end{tabular}
		}
	\end{center}
\end{table*}

\begin{table*}
	\centering
	\caption{We report the performances of HME, HCRN, and PSAC on the RWR-n metrics (the lower the better), where n represents the number of incorrectly answered sub-questions for a composition, conditioned on the given composition type.}
	\label{tab_9_combination_RWR}
	\begin{center}
		\scalebox{0.83}{
			\begin{tabular}{lrrrrrrrrrrrr}
				\multirow{2}{*}{ Composition Type } & \multicolumn{3}{c}{ HME } & \multicolumn{3}{c}{ HCRN } & \multicolumn{3}{c}{ PSAC } & \multicolumn{3}{c}{ Ours } \\
				\cmidrule(r){2-4} \cmidrule(r){5-7}  \cmidrule(r){8-10} \cmidrule(r){11-13}
				& RWR-1 & RWR-2 & RWR-3 & RWR-1 & RWR-2 & RWR-3 & RWR-1 & RWR-2 & RWR-3 & RWR-1 & RWR-2 & RWR-3 \\
				\hline \hline 
				Interaction & 60.96 & 89.01 & 9.84 & 59.02 & 58.15 & 25.37 & 45.50 & 28.86 & 29.23 & 39.11 & 20.36 & 11.79 \\
				 First & 8.62 & N/A & N/A & 0.01 & N/A & N/A & 7.90 & N/A & N/A & 1.72 & N/A & N/A \\
				Last & 11.93 & N/A & N/A & 0.01 & N/A & N/A & 7.65 & N/A & N/A & 6.36 & N/A & N/A \\
				Action & 1.59 & N/A & N/A & 7.69 & N/A & N/A & 2.50 & N/A & N/A & 66.85 & N/A & N/A \\
				Object & 40.33 & 22.12 & N/A & 33.06 & 31.76 & N/A & 30.46 & 24.53 & N/A & 8.89 & 16.65 & N/A \\
				Equals & 58.21 & 49.10 & N/A & 50.64 & 50.62 & N/A & 36.01 & 20.78 & N/A & 50.54 & 45.29 & N/A \\
				 And & 100 & 0.00 & N/A & 32.99 & 17.22 & N/A & 18.70 & 5.07 & N/A & 40.94 & 29.01 & N/A \\
				Xor & 14.98 & 100 & N/A & 61.47 & 31.93 & N/A & 76.07 & 7.19 & N/A & 16.88 & 98.03 & N/A \\
				 Choose & 44.35 & 81.48 & N/A & 45.08 & 45.10 & N/A & 55.60 & 32.46 & N/A & 29.26 & 55.62 & N/A \\
				Longer Choose & 50.00 & 19.17 & N/A & 34.18 & 32.53 & N/A & N/A & N/A & N/A & N/A & N/A & N/A \\
				Shorter	Choose & 0.00 & 18.27 & N/A & 31.35 & 35.26 & N/A & N/A & N/A & N/A & N/A & N/A & N/A \\
				After & 60.95 & 19.80 & N/A & 61.40 & 31.61 & N/A & 24.92 & 10.11 & N/A & 22.69 & 17.50 & N/A \\
				Before & 60.49 & 19.18 & N/A & 59.64 & 29.78 & N/A & 24.98 & 7.58 & N/A & 23.30 & 21.03 & N/A \\
				While & 64.97 & 6.63 & N/A & 64.41 & 19.14 & N/A & 27.06 & 8.43 & N/A & 9.48 & 6.39 & N/A \\
				Between & 45.79 & 1.31 & N/A & 36.44 & 6.36 & N/A & 35.54 & 4.68 & N/A & 7.63 & 17.07 & N/A \\
				\hline Overall & 45.42 & 22.96 & 9.84 & 49.05 & 35.55 & 25.37 & 36.24 & 20.32 & 29.23 & 26.45 & 20.14 & 11.79 \\
			\end{tabular}
		}
	\end{center}
\end{table*}

\section{Metrics}
\label{sec:Metrics}
The performances were evaluated using the metrics CA, RWR, Delta, and IC, and Accuracy proposed in the AGQA-Decomp benchmark. The definitions of these metrics are described above.

\begin{table}
	\centering
	\caption{We present internal consistency (IC) scores for individual logical consistency rules for HCRN, HME, PSAC and the Ours model. The N/A indicates that the amount of data for a combination type is 0.}
	\label{tab_10_question_yesno}
	\begin{center}
		\scalebox{0.75}{
			\begin{tabular}{lcrrrr}
				\multirow{2}{*}{Consistency Check}  & \multirow{2}{*}{Parent Answer} & \multicolumn{4}{c}{$\mathrm{IC} \uparrow$} \\
				\cline{3-6} & & HME & HCRN & PSAC & Ours \\
				 \hline \hline \multirow{2}{*}{Interaction } & Yes & N/A & 50.62 & 79.39 & 24.17 \\
				& No & 100 & 46.6 & N/A & 98.11 \\
				\hline \multirow{2}{*}{First } & Object & 38.23 & 0.02 & 74.95 & 33.64 \\
				 & Action & 11.83 & 1.08 & 76.52 & 70.61 \\
				\hline \multirow{2}{*}{Last } & Object & 35.81 & 0.01 & 76.11 & 19.18 \\
				& Action & 7.58 & 0.76 & 87.12 & 25.76 \\
				\hline \multirow{2}{*}{Action } & After & 98.15 & 57.41 & 21.37 & 44.30 \\
				& Before & 98.42 & 47.89 & 8.95 & 87.89 \\
				\hline \multirow{4}{*}{Object } & After & 77.35 & 55.55 & 42.25 & 35.55 \\
				& Before & 76.96 & 56.94 & 43.26 & 33.44 \\
				& While &78.35 & 58.99 & 43.87 & 35.01 \\
				& Between & 75.49 & 55.37 & 29.42 & 30.59 \\
				\hline \multirow{2}{*}{Equals } & Yes & 0.00 & 5.18 & 2.72 & 5.45 \\
				& No & 97.80 & 36.37 & 0.00 & 88.41 \\
				\hline \multirow{2}{*}{And } & Yes & N/A & 73.58 & 78.26 & 31.80 \\
				& No & 100 & 49.99 & N/A & 70.23 \\
				\hline \multirow{2}{*}{Xor } & Yes & N/A & 8.95 & 0.00 & 66.51 \\
				& No & 100 & 22.33 & 0.00  & 97.85 \\
				\hline \multirow{3}{*}{Choose } & Temporal & 0.00 & 0.47 & 0.00 & 0.86 \\
				& Object & 0.00 & 11.54 & 0.00 & 96.52 \\
				& Action & 0.00 & 29.11 & N/A & 50.19 \\
				\hline \multirow{2}{*}{After } & Yes & N/A & 48.78 & 2.08 & 76.29 \\
				& No & 99.97 & 46.70 & 56.52 & 33.26 \\
				\hline \multirow{2}{*}{Before } & Yes & N/A & 50.64 & 1.98 & 78.37 \\
				& No & 99.97 & 52.07 & 56.89 & 34.01 \\
				\hline \multirow{2}{*}{While } & Yes & N/A  & 52.28 & 2.30 & 77.71 \\
				& No & 100 & 53.43 & 70.18 & 13.52 \\
				\hline \multirow{2}{*}{Between } & Yes & N/A  & 81.10 & 88.13 & 38.65 \\
				& No & 100 & 68.12 & 99.73 & 18.25 \\
				\hline \multirow{2}{*}{Object } & Yes & 0.00 & 35.73 & 100 & 100 \\
				& No & 100 & 66.95 & 0 & 69.69 \\
				\hline Overall & - & 87.32 & 44.54 & 59.49 & 42.81\\
			\end{tabular}
		}
	\end{center}
\end{table}

\begin{table}
	\centering
	\caption{We report accuracy per ground-truth answer for each binary sub-question type expecting ``Yes” or ``No” answers for HCRN, HME, PSAC and Ours model. ``-” indicates there were no results for a given type, N/A indicates that the amount of data for a question type is 0. HME particularly is biased towards ``No” for all question types, while PSAC particularly is biased towards ``Yes” for all question types.}
	\label{tab_11_combination_yesno}
	\begin{center}
		\scalebox{0.70}{
			\begin{tabular}{lcrrrr}
				\multirow{2}{*}{Sub-question Type}  & \multirow{2}{*}{Ground Truth} & \multicolumn{4}{c}{$\mathrm{ACC} \uparrow$} \\
				\cline{3-6} & & HME & HCRN & PSAC & Ours \\
				\hline \hline \multirow{2}{*}{Object Exists } & Yes & 0.00 & 31.69 & 88.10 & 96.26 \\
				 & No & N/A & N/A & N/A & N/A \\
				\hline \multirow{2}{*}{Relation Exists } & Yes & 0.00 & 61.72 & 100 & 72.94 \\
				 & No & N/A & N/A & N/A & N/A \\
				\hline \multirow{2}{*}{Interaction } & Yes & 0.00 & 55.11 & 83.75& 57.20 \\
				 & No & 100 & 45.37 & 0.00 & 86.30 \\
				\hline \multirow{2}{*}{Exists Temporal Loc. } & Yes & 0.00 & 64.01 & 79.10 & - \\
				 & No & 100 & 26.54 & 0.00 & - \\
				\hline \multirow{2}{*}{Interaction Temporal Loc.} & Yes & 0.00 & 28.12 & 78.73 & 21.64 \\
				 & No & 100 & 73.58 & 0.01 & 79.99 \\
				\hline \multirow{2}{*}{Conjunction} & Yes & 0.00 & 24.34 & 89.61 & 96.28 \\
				& No & 100 & 75.96 & 0.00 & 5.85 \\
				\hline \multirow{2}{*}{Equals } & Yes & 0.00 & 61.64 & 62.00 & 10.96 \\
				& No & 100 & 38.72 & 0.00 & 95.12 \\
			\end{tabular}
		}
	\end{center}
\end{table}
\textbf{Notations.} Here, we denote each test sample in the test set $S$ as $(v,q)$, where $v$ and $q$ refer to video and question respectively. Let $gt(v,q)$ be the ground-truth answer of $(v,q)$, and $f(v,q)$ be the answer predicted by model $f$. For convenience, we define the video-question pairs correctly answered by $f$ in $S$ as $a( S,f)$, formally, $a(S, f)=\{(v, q) \mid(v, q) \in S, f(v, q)=g t(v, q)\}$. Additionally, for compositionality analysis, we denote the set of compositional questions as $Q_c$, where each $q \in Q_c$ has a set of sub-questions $C(q)$.

\textbf{Compositional accuracy (CA).} CA tests if the model can answer a question correctly based on correct intermediate results. Let $S_{\mathrm{CA}}$ be the set of video-question pairs that all direct sub-questions have answered correctly by $f$, i.e., $S_{\mathrm{CA}} \!=\! \left\{(v, q)\left|(v, q) \in S, q \in Q_c,\right| a(C(q), f)|=| C(q) \mid\right\}$. The CA is calculated by:
\begin{small}
	\begin{equation}
		\begin{gathered}
			CA(f)=\frac{\left|a\left(S_{\mathrm{CA}}, f\right)\right|}{\left|S_{\mathrm{CA}}\right|}  \label{con:eq_21}
		\end{gathered}
	\end{equation}
\end{small}

\textbf{Right for the wrong reasons (RWR).} RWR checks if the model guessed the answer correctly but based on wrong intermediate results. Let $S_\mathrm{RWR}$ be the set of video-question pairs that has some direct sub-question answered incorrectly by $f$, i.e., $S_\mathrm{RWR}=Q_c-S_\mathrm{CA}$. The RWR is calculated by:
\begin{small}
	\begin{equation}
		\begin{gathered}
			RWR(f)=\frac{\left|a\left(S_{\mathrm{RWR}}, f\right)\right|}{\left|S_{\mathrm{RWR}}\right|}  \label{con:eq_22}
		\end{gathered}
	\end{equation}
\end{small}

In addition, following \cite{gandhi2022measuring}, we measure the detailed RWR of our NS-VideoQA by the RWR-$n$ metric, where $n$ represents the exact number of incorrectly answered sub-questions for a composition, i.e., to replace $S_\mathrm{RWR}$ with $S_{\mathrm{RWR}-n} \!=\! \left\{(v, q)\left|(v, q) \in S, q \in Q_c,\right| a(C(q), f)|=| C(q) \mid-n\right\}$.

\textbf{Delta.} Delta represents the discrepancy between RWR and CA. The Delta is calculated by:
\begin{small}
	\begin{equation}
		\begin{gathered}
			Delta(f)=RWR(f)-CA(f)  \label{con:eq_23}
		\end{gathered}
	\end{equation}
\end{small}

A positive Delta value indicates that for the model $f$, answering sub-questions incorrectly leads to higher accuracy, which reveals the irrational reasoning steps taken by the model $f$.

\textbf{Internal Consistency (IC).} IC quantifies the ability to avoid self-contradictions. To measure IC, a set of logical consistency rules $\Phi$ is given in AGQA-Decomp. For each rule $\phi \!\in\! \Phi$, each question $q$ derives $R_\phi(q)$, a set of questions such that for $\forall q^{\prime} \!\in\! R_\phi(q)$, the answer to $q^{\prime}$ can be inferred from the answer to $q$ according to the rule $\phi$. Then, the IC can be calculated by:
\begin{small}
	\begin{equation}
		\begin{gathered}
			IC(f)=\frac{\sum_{\phi \in \Phi} IC_\phi(f)}{|\Phi|} \label{con:eq_24}
		\end{gathered}
	\end{equation}
\end{small}
\begin{small}
	\begin{equation}
		\begin{gathered}
				IC_\phi(f)=\frac{\sum_{(v, q) \in S, q \in Q_c}\left|a\left(R_\phi(q), f\right)\right|}{\sum_{(v, q) \in S, q \in Q_c}\left|R_\phi(q)\right|}  \label{con:eq_25}
		\end{gathered}
	\end{equation}
\end{small}

\textbf{Accuracy.} The Accuracy scores are calculated for each question type, and the $Overall$ accuracy is the average accuracy of all question types. Formally, let $T$ be the set of all question types. Each question type $t \!\in\! T$ has an answer set $A_t$. The questions of type $t$ with a specific ground truth answer $g \!\in\! A_t$ are denoted as $Q_{t, g} \!\subsetneq\! S$. Then, the accuracy score $Accuracy(f,t)$ for model $f$ and question type $t$ is calculated by:
\begin{small}
	\begin{equation}
		\begin{gathered}
			Accuracy(f, t)=\frac{1}{\left|A_t\right|} \sum_{g \in A_t}\left(\frac{1}{\left|Q_{t, g}\right|} a\left(Q_{t, g}, f\right)\right)\label{con:eq_26}
		\end{gathered}
	\end{equation}
\end{small}

The overall accuracy is:
\begin{small}
	\begin{equation}
		\begin{gathered}
				Accuracy_{all}(f)=\frac{1}{|T|} \sum_{t \in T} Accuracy(f, t)  \label{con:eq_27}
		\end{gathered}
	\end{equation}
\end{small}

\section{Performance on composition questions}
\label{sec:Performance}
In the experimental section sec.~\ref{sec:Experiments}, we reported the evaluation results of Accuracy, CA, RWR, Delta, and IC on different sub-questions. The experimental results demonstrate that the NS-VideoQA model possesses a certain level of compositional reasoning ability and meets the internal consistency check. To further validate the compositional reasoning capability of NS-VideoQA, we conducted supplemental experiments on composite questions of AGQA Decomp. The distribution of combination types is illustrated in Figure~\ref{distribution_combination}.

Based on the experimental results presented in Tables~\ref{tab_8_combination_CA},~\ref{tab_9_combination_RWR},~\ref{tab_10_question_yesno},~\ref{tab_11_combination_yesno}, it can be observed that the HME \cite{fan2019heterogeneous} model is biased towards ``No" answers, while the PSAC \cite{li2019beyond}model is biased towards ``Yes" answer. Thereforce, these models exhibit limitations in their ability to reason over given compositions, often resulting in correct answers for the wrong reasons, which can be attributed to self-contradiction. To illustrate, let's consider the $Interaction$ composition. The HME model achieves the following values in Table~\ref{tab_8_combination_CA}: CA ($100\%$), RWR ($36.71\%$), Delta (-$63.29\%$), and IC ($50\%$). In Table~\ref{tab_9_combination_RWR}, the model achieves RWR-1 ($60.96\%$), RWR-2 ($89.01\%$), and RWR-3 ($9.84\%$). Moving on to Table~\ref{tab_10_question_yesno}, we find the IC values for $Interaction$-Yes and $Interaction$-No as (N/A, $100\%$). Here, N/A indicates that there are no instances where $Interaction$ compositions result in ``Yes" answers. Furthermore, Table~\ref{tab_11_combination_yesno} shows that the Accuracy values for $Interaction$-Yes and $Interaction$-No are ($0.00\%$, $100\%$). This is sufficient to prove the above conclusion.

Therefore, let's briefly compare the differences between HCRN \cite{le2020hierarchical} and NS-VideoQA in terms of experimental results. Our model demonstrates a stronger advantage in compositional reasoning tasks. For the $First$, $Last$, $Action$, $Object$, and $Choose$ compositions, our model surpasses HCRN in terms of CA values by $21.31\%$, $92.61\%$, $70.85\%$, -$1.99\%$, and $46.24\%$ respectively. Similarly, our model outperforms HCRN in IC values by $12.53\%$, $52.09\%$, $22.45\%$, $1.67\%$, and $35.79\%$ respectively. Regarding the $And$ and $Xor$ compositions, our model excels in answering $Xor$, while HCRN performs better in answering $And$. In terms of $After$, $Before$, $While$, and $Between$ compositions, the overall differences between our model and HCRN, as observed from Tables~\ref{tab_8_combination_CA},~\ref{tab_9_combination_RWR},~\ref{tab_10_question_yesno},~\ref{tab_11_combination_yesno}, are not significant. Although HCRN achieves slightly higher CA and IC values compared to our model, our model exhibits relatively lower RWR values, with reductions of $27.66\%$, $27.17\%$, $39.71\%$, and $2.78\%$ respectively. Additionally, our model shows significantly lower RWR-1 and RWR-2 values compared to HCRN, indicating that HCRN relies more on guessing the answers to compositional questions based on incorrect sub-question results. For $Overall$, our model surpasses HCRN by $2.19\%$ in terms of IC values.

\section{Extra Examples from AGQA Decomp}
\label{sec:Examples}
\begin{figure*}[htbp]
	\centering
	\setlength{\belowcaptionskip}{-0.4cm}
	\includegraphics[width=170mm]{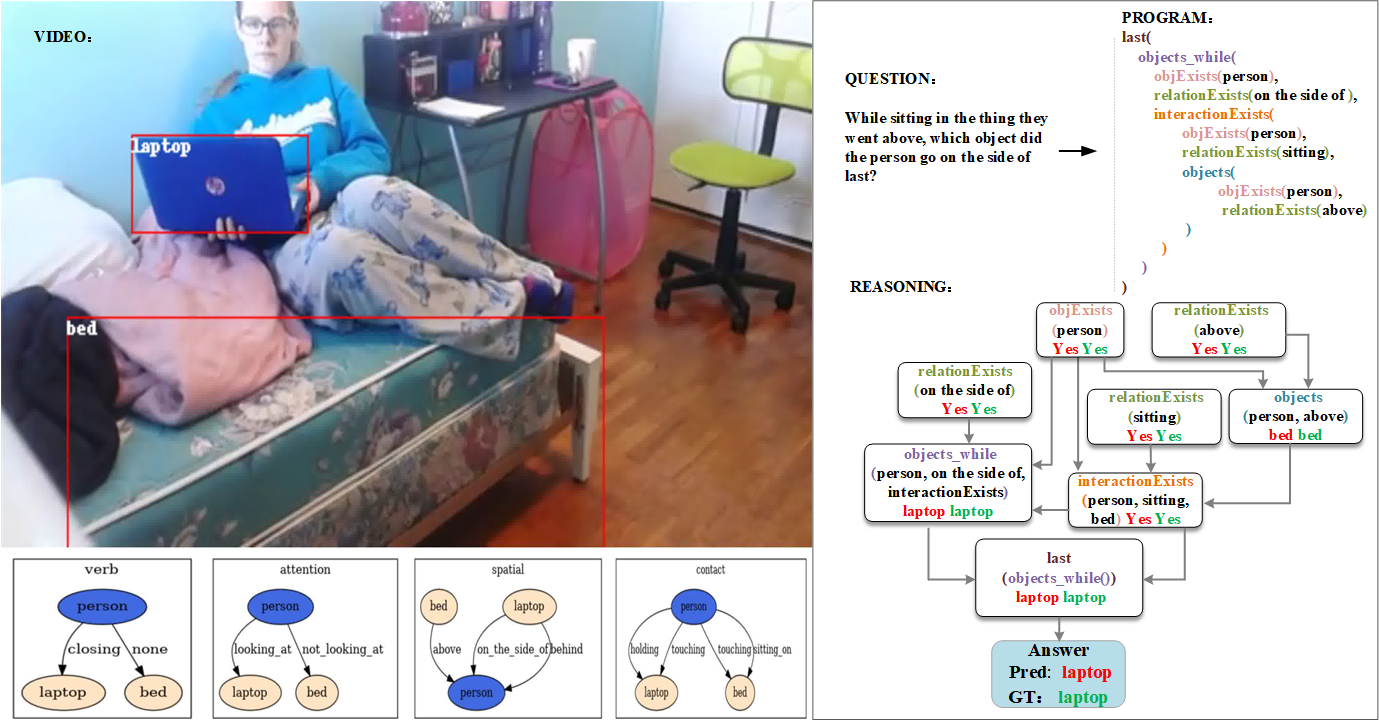}
	\caption{An example of a type $Last$ inference process, where red indicates the prediction from NS-VideoQA and green represents the Ground Truth (GT).
	}
	\label{picture-7}
\end{figure*}

\begin{figure*}[htbp]
	\centering
	\setlength{\belowcaptionskip}{-0.4cm}
	\includegraphics[width=170mm]{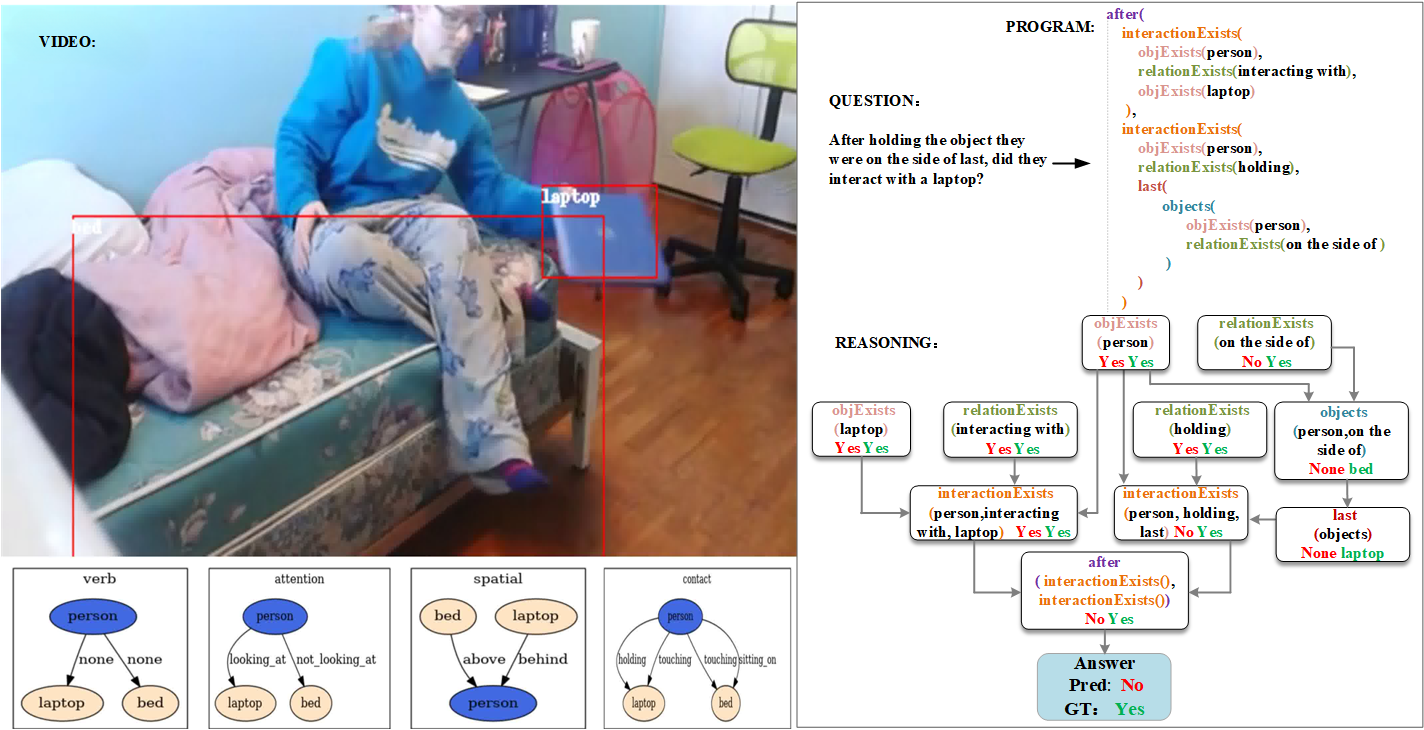}
	\caption{An example of a type $Interaction\ Temporal\ Loc.$ inference process, where red indicates the prediction from NS-VideoQA and green represents the Ground Truth (GT).
	}
	\label{picture-8}
\end{figure*}

\begin{figure*}[htbp]
	\centering
	\setlength{\belowcaptionskip}{-0.4cm}
	
	\includegraphics[width=168mm]{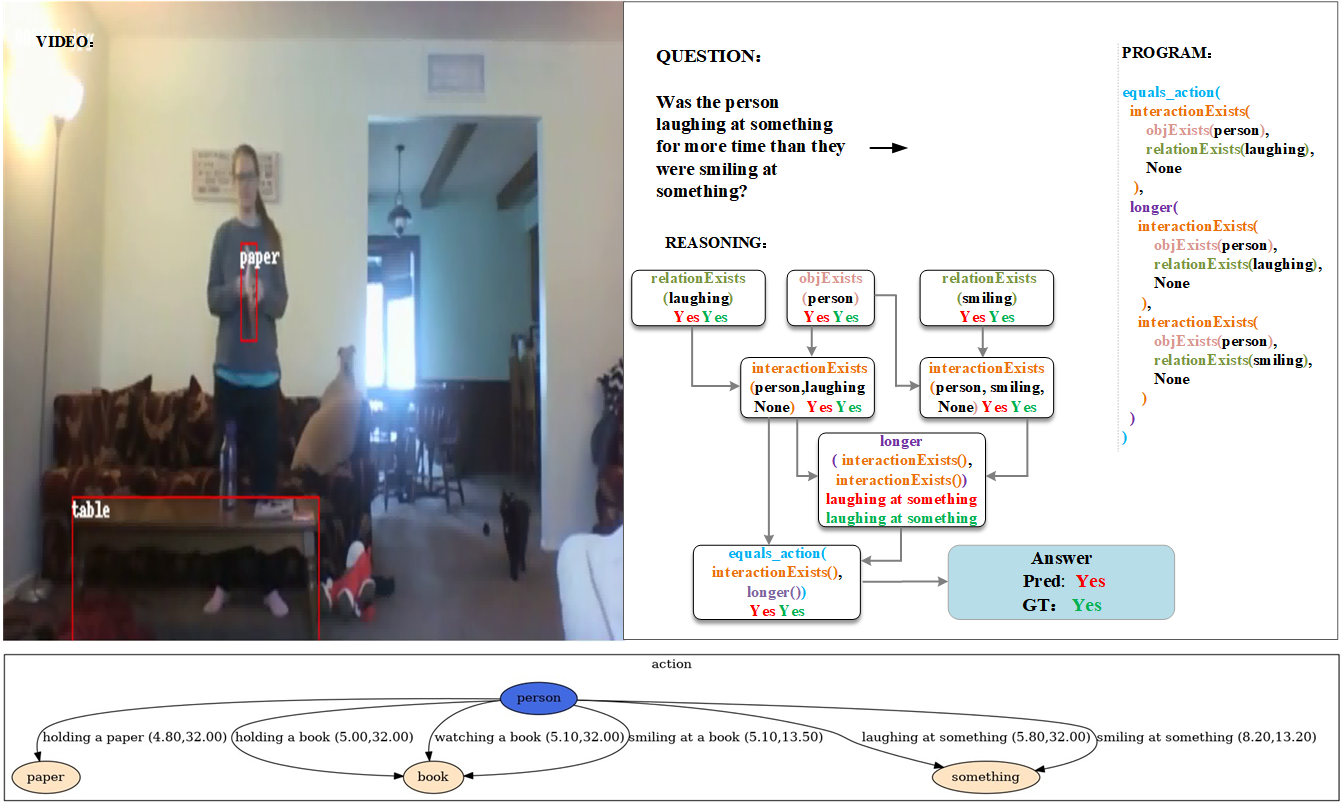}
	\caption{An example of a type $Equals$ inference process, where red indicates the prediction from NS-VideoQA and green represents the Ground Truth (GT).
	}
	\label{picture-9}
\end{figure*}

\begin{figure*}[htbp]
	\centering
	\setlength{\belowcaptionskip}{-0.4cm}
	
	\includegraphics[width=168mm]{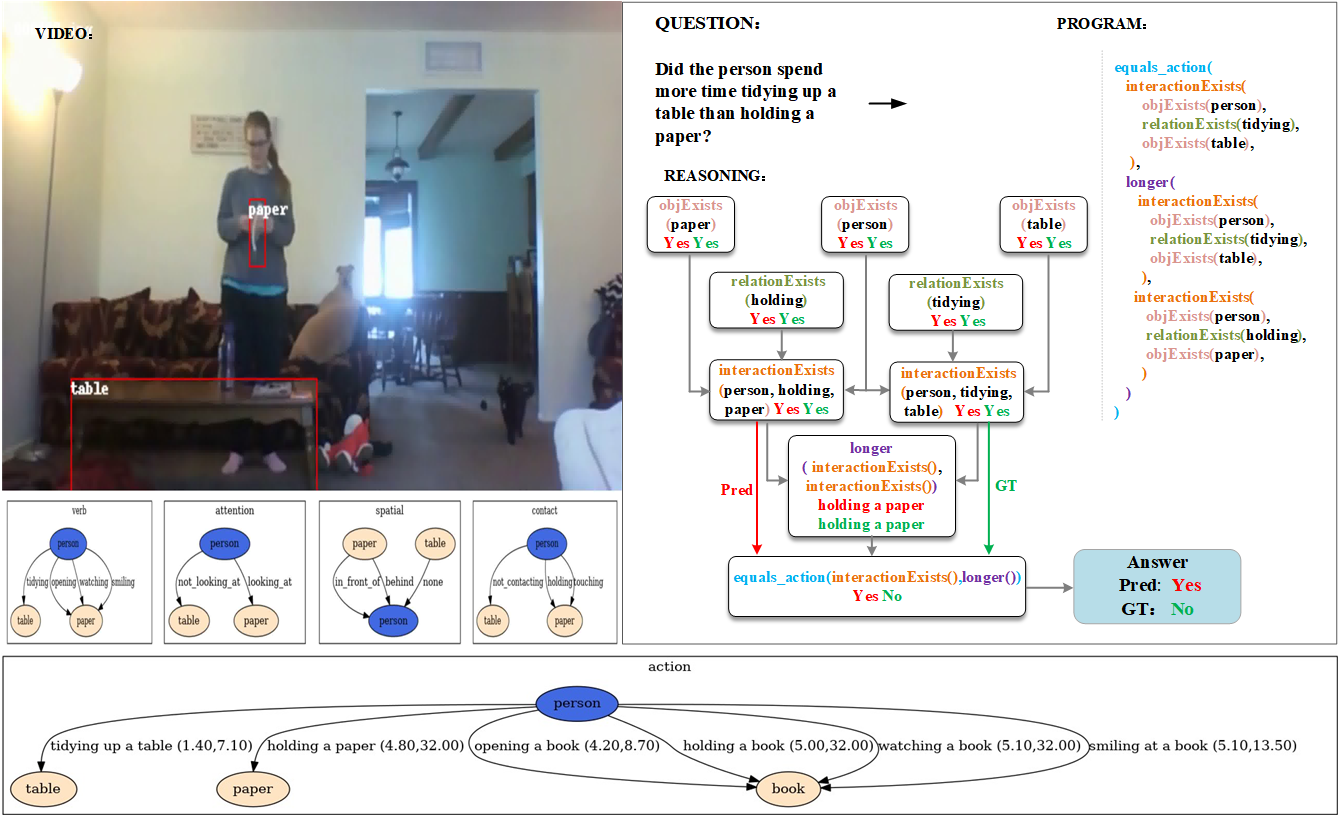}
	\caption{An example of a type $Equals$ inference process, where red indicates the prediction from NS-VideoQA and green represents the Ground Truth (GT). The generation order of the two $interactionExists$ sub-questions is swapped, resulting in passing incorrect parameters to the equals function (as indicated by the red arrow), leading to the opposite result.
	}
	\label{picture-10}
\end{figure*}

\begin{figure*}[htbp]
	\centering
	\setlength{\belowcaptionskip}{-0.4cm}
	
	\includegraphics[width=170mm]{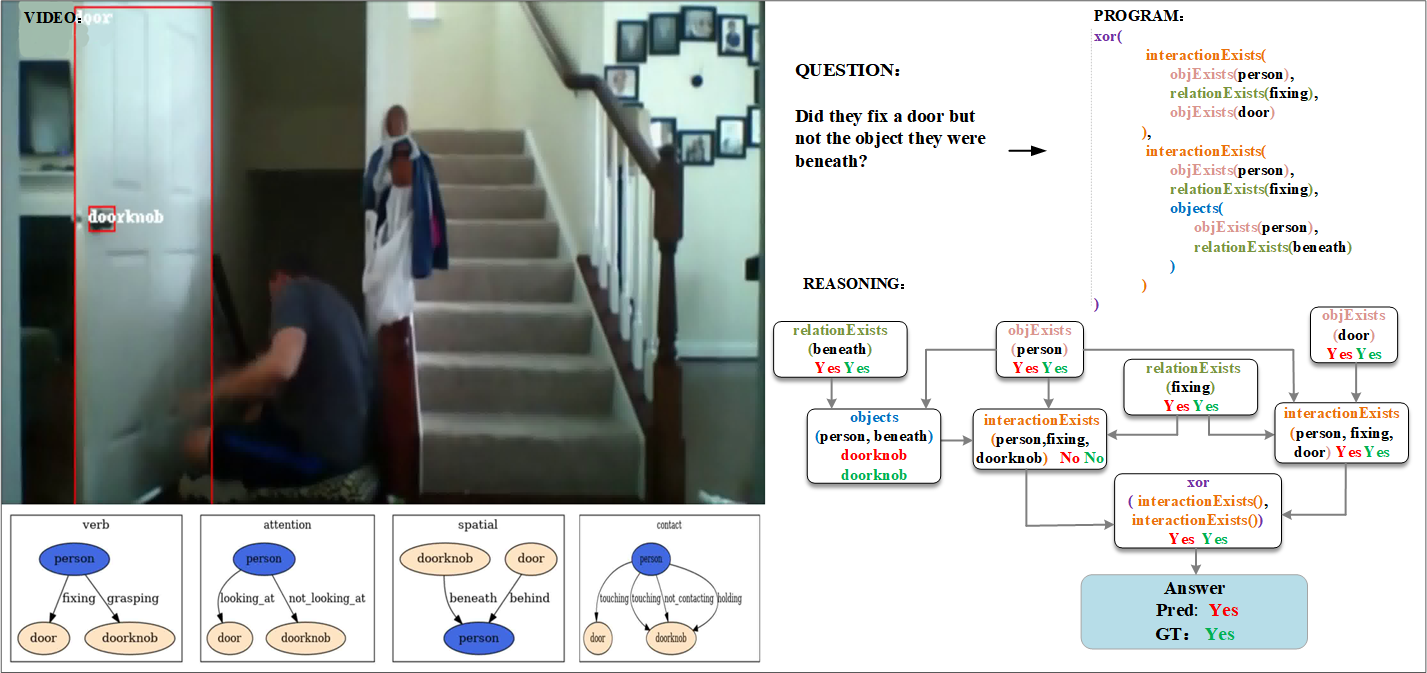}
	\caption{An example of a type $Conjunction$ inference process, where red indicates the prediction from NS-VideoQA and green represents the Ground Truth (GT).}
	\label{picture-11}
\end{figure*}

\begin{figure*}[htbp]
	\centering
	\setlength{\belowcaptionskip}{-0.4cm}
	
	\includegraphics[width=170mm]{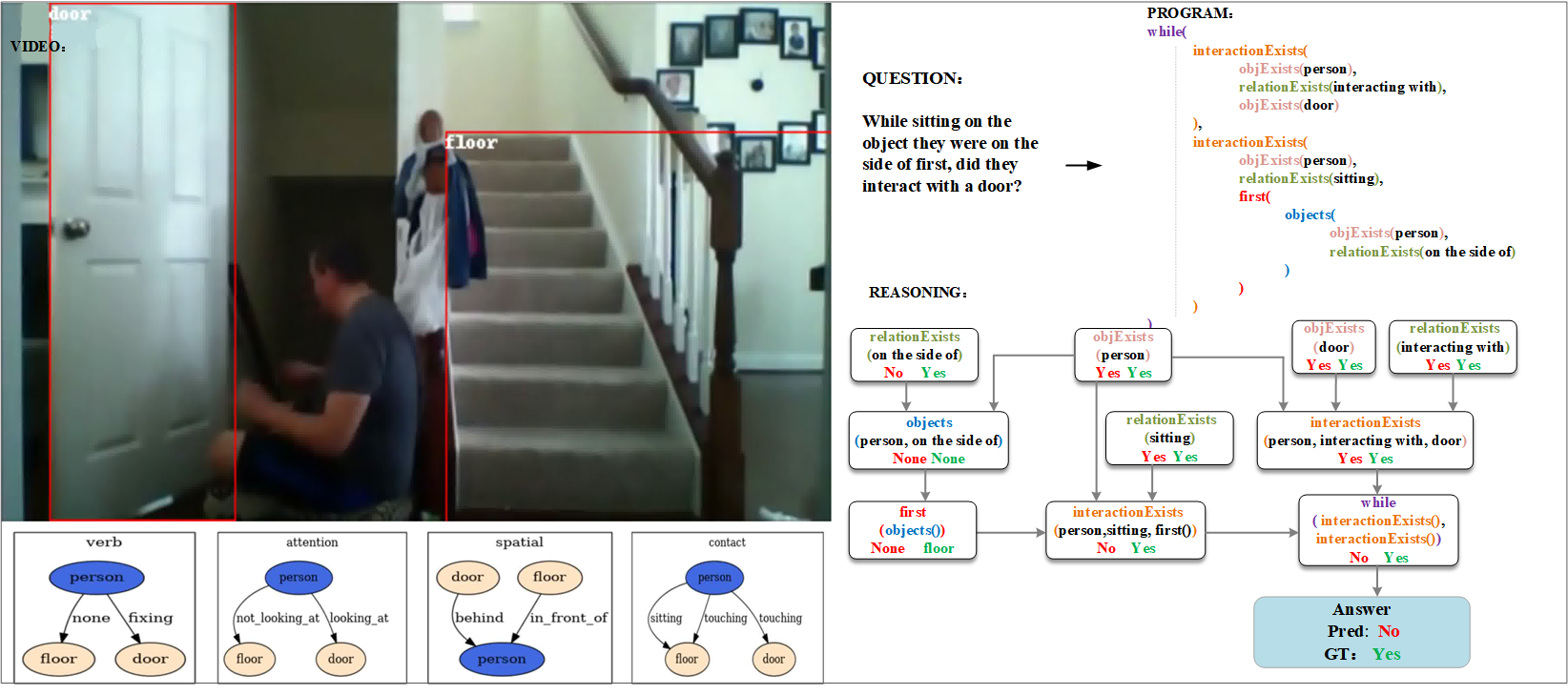}
	\caption{An example of a type $Interaction\ Temporal\ Loc.$ inference process, where red indicates the prediction from NS-VideoQA and green represents the Ground Truth (GT).
	}
	\label{picture-12}
\end{figure*}

\begin{figure*}[htbp]
	\centering
	\setlength{\belowcaptionskip}{-0.4cm}
	
	\includegraphics[width=170mm]{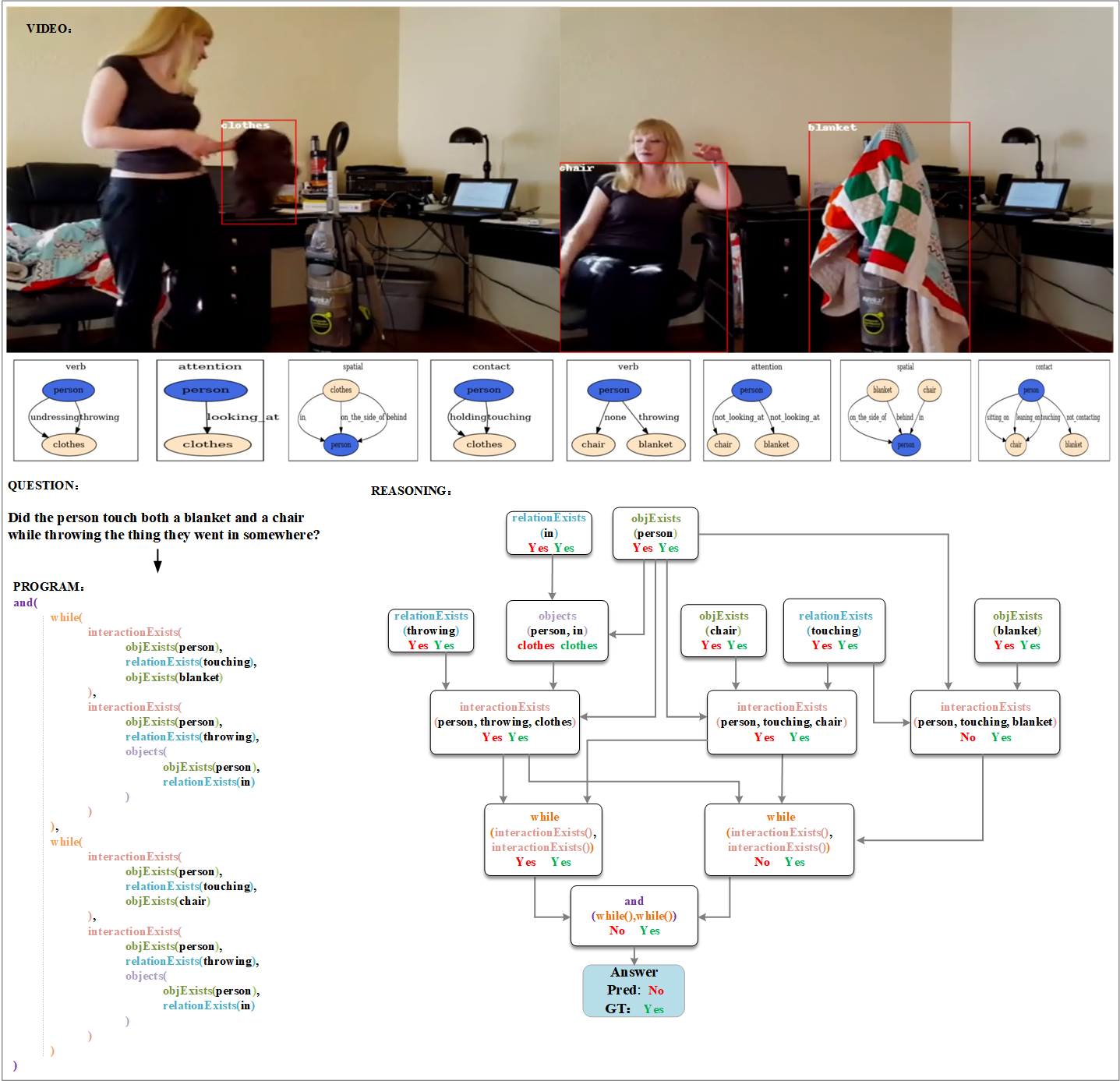}
	\caption{An example of a type $Conjunction$ inference process, where red indicates the prediction from NS-VideoQA and green represents the Ground Truth (GT).
	}
	\label{picture-13}
\end{figure*}

\begin{figure*}[htbp]
	\centering
	\setlength{\belowcaptionskip}{-0.4cm}
	
	\includegraphics[width=170mm]{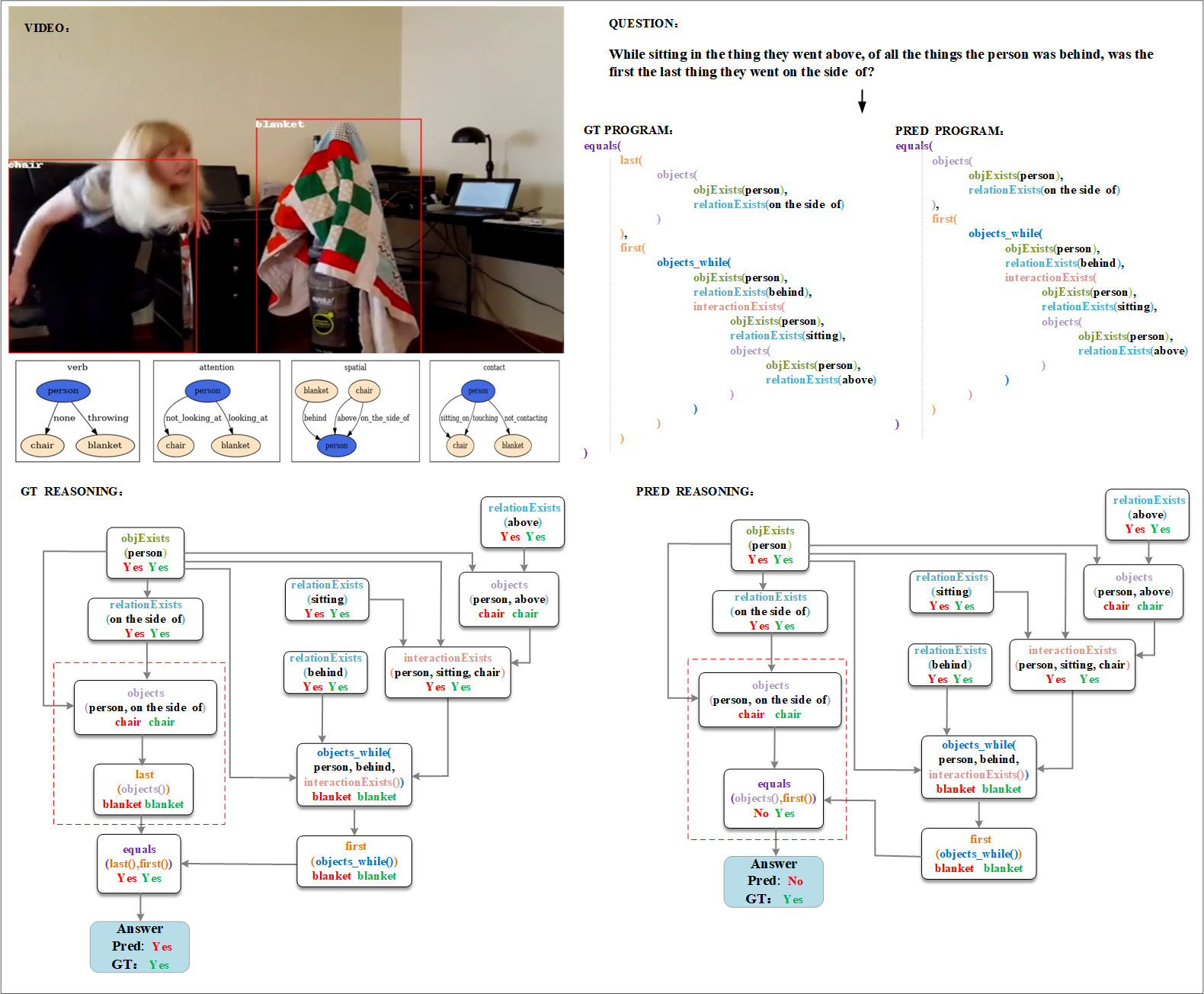}
	\caption{An example of a type $Equals$ inference process, where red indicates the prediction from NS-VideoQA and green represents the Ground Truth (GT). The Language Question Parser incorrectly predicts the program equals (objects (), first()) (marked by the red dotted line in the diagram), leading to a failed final result.
	}
	\label{picture-14}
\end{figure*}

In this section, we will showcase the spatio-temporal reasoning process for a variety of question types in Figure~\ref{picture-7},~\ref{picture-8},~\ref{picture-9},~\ref{picture-10},~\ref{picture-11},~\ref{picture-12},~\ref{picture-13},~\ref{picture-14}, such as $Interaction\ Temporal\ Loc.$, $Conjunction$, $First/Last$ and $Equals$ categories. These examples will include instances of correct reasoning and instances of erroneous reasoning in NS-VideoQA. In the incorrect examples, the differences in symbolic reasoning programs stem from the variations in the descriptions of the input natural language questions. Our Language Question Parser, which utilizes a Bi-LSTM Encoder-Decoder, remains uninterpretable internally. Fortunately, in the NS-VideoQA framework, the neural-symbolic integration architecture allows us to conduct error analysis around the reasoning process and pinpoint the issue to the Language Question Parser, inspiring future improvements in program generation methods.


\begin{table*}
	\centering
	\caption{Reasoning rules corresponding to $Object\ Exists$, $Relation\  Exists$, and $Interaction$ question.}
	\label{tab12_rules_easy}
	\begin{center}
		\scalebox{1}{
			\begin{tabular}{|c|c|c|c|}
				\hline \begin{tabular}{c} 
					Sub-question Type
				\end{tabular} & Reasoning Rule/ Description & Input Type & Output Type \\
				\hline
				\multirow{4}{*}[-1.5ex]{$Object\ Exists$}
				& \textbf{filter\_object} & \multirow{2}{*}[-1.5ex]{ (scene, token) } & \multirow{2}{*}[-1.5ex]{ objects} \\
				\cline{2-2} & \begin{tabular}{l} 
					Selects objects from the input scene list \\
					with the input object.
				\end{tabular} & & \\
				\cline{2-4} & \textbf{query\_object} & \multirow{2}{*}{ (exe\_trace, token) } & \multirow{2}{*}{ binary } \\
				\cline{2-2} & To verify if an object exists from objects. & & \\
				\hline \multirow{4}{*}[-1.5ex]{ $Relation\ Exists$ } &\textbf{filter\_relation} & \multirow{2}{*}[-1.5ex]{ (scene, token) } & \multirow{2}{*}[-1.5ex]{ relations } \\
				\cline{2-2} & \begin{tabular}{l} 
					Selects relations from the input scene list \\
					with the input relation.
				\end{tabular} & & \\
				\cline{2-4} & \textbf{query\_relation} & \multirow{2}{*}{ (exe\_trace, token) } & \multirow{2}{*}{ binary } \\
				\cline{2-2} & To verify if a relation exists from relations. & & \\
				\hline \multirow{2}{*}{ $Interaction$ } & \textbf{query\_interaction} & \multirow{2}{*}{ (exe\_trace, token) } & \multirow{2}{*}{ binary } \\
				\cline{2-2} & To verify if an interaction exists. & & \\
				\hline
			\end{tabular}
		}
	\end{center}
\end{table*}

\section{Reasoning rules of SRM}
\label{sec:Reasoning}
In this section, we present the definitions of reasoning rules for each sub-question type, including functional descriptions and input/output specifications. Specifically, Tables~\ref{tab12_rules_easy},~\ref{tab13_rules_middle},~\ref{tab14_rules_difficult} list the reasoning rules corresponding to all sub-questions, while Table~\ref{tab15_rules_in/out} provides the definitions of input/output data types for the reasoning rules.
\begin{table*}
	\centering
	\caption{Reasoning rules corresponding to $Temporal\ loc.$, $Longest/Shortest\ Action$, $Action$, and $Object$ question.}
	\label{tab13_rules_middle}
	\begin{center}
		\scalebox{1}{
			\begin{tabular}{|c|c|c|c|}
				\hline \begin{tabular}{c} 
					Sub-question Type
				\end{tabular} & Reasoning Rule/ Description & Input Type & Output Type \\
				\hline
				\multirow{8}{*}[-8.5ex]{\begin{tabular}{l} 
						$Interaction$ \\
						$Temporal\ loc.$/ \\
						$Exists$ \\
						$Temporal\ loc.$
				\end{tabular}} & \textbf{interaction\_temporal\_after} & \multirow{2}{*}[-2.5ex]{ (exe\_trace, token) } & \multirow{2}{*}[-2.5ex]{ binary } \\
				\cline{2-2} & \begin{tabular}{l} 
					Combine two interaction or exists questions \\
					 using a temporal localizer ``after".
				\end{tabular} & & \\
				\cline{2-4} & \textbf{interaction\_temporal\_before} & \multirow{2}{*}[-2.5ex]{ (exe\_trace, token) } & \multirow{2}{*}[-2.5ex]{ binary } \\
				\cline{2-2} & \begin{tabular}{l} 
					Combine two interaction or exists questions \\
					 using a temporal localizer ``before".
				\end{tabular} & & \\
				\cline{2-4} & \textbf{interaction\_temporal\_while} & \multirow{2}{*}[-1.5ex]{ (exe\_trace, token) } & \multirow{2}{*}[-1.5ex]{ binary } \\
				\cline{2-2} & \begin{tabular}{l} 
					Combine two interaction or exists questions \\
					 using a temporal localizer ``while".
				\end{tabular} & & \\
				\cline{2-4} & \textbf{interaction\_temporal\_between} & \multirow{2}{*}[-2.5ex]{ (exe\_trace, token) } & \multirow{2}{*}[-2.5ex]{ binary } \\
				\cline{2-2} & \begin{tabular}{l} 
					Combine two interaction or exists questions  \\
					using a temporal localizer ``between".
				\end{tabular} & & \\
				
				\hline
				\multirow{4}{*}[-2.5ex]{\begin{tabular}{l} 
						$Action$ \\
						$Temporal\ loc.$
				\end{tabular}} & \textbf{actions\_after} & \multirow{2}{*}[-1.5ex]{ (exe\_trace, token) } & \multirow{2}{*}[-1.5ex]{ action } \\
				\cline{2-2} & \begin{tabular}{l} 
					Getting the action that occurs after an \\ interaction.
				\end{tabular} & & \\
				\cline{2-4} & \textbf{actions\_before} & 
				\multirow{2}{*}[-1.5ex]{ (exe\_trace, token) } & \multirow{2}{*}[-1.5ex]{ action } \\
				\cline{2-2} & \begin{tabular}{l} 
					Getting the action that occurs before an \\ interaction.
				\end{tabular} & & \\
				
				\hline
				\multirow{8}{*}[-10.5ex]{\begin{tabular}{l} 
						$Object$ \\
						$Temporal\ loc.$
				\end{tabular}} & \textbf{objects\_after} & \multirow{2}{*}[-2.5ex]{ (exe\_trace, token) } & \multirow{2}{*}[-2.5ex]{ object } \\
				\cline{2-2} & \begin{tabular}{l} 
					Selects objects that involved with the specified  \\
					interaction appears after the specified action.
				\end{tabular} & & \\
				\cline{2-4} & \textbf{objects\_before} & \multirow{2}{*}[-2.5ex]{ (exe\_trace, token) } & \multirow{2}{*}[-2.5ex]{ object } \\
				\cline{2-2} & \begin{tabular}{l} 
					Selects objects that involved with the specified \\ 
					interaction appears before the specified action.
				\end{tabular} & & \\
				\cline{2-4} & \textbf{objects\_while} & \multirow{2}{*}[-1.5ex]{ (exe\_trace, token) } & \multirow{2}{*}[-1.5ex]{ object } \\
				\cline{2-2} & \begin{tabular}{l} 
					Selects objects that involved with the specified \\
					interaction appears while the specified action.
				\end{tabular} & & \\
				\cline{2-4} & \textbf{objects\_between} & \multirow{2}{*}[-2.5ex]{ (exe\_trace, token) } & \multirow{2}{*}[-2.5ex]{ object } \\
				\cline{2-2} & \begin{tabular}{l} 
					Selects objects that involved with the specified \\
					interaction appears between the two specified\\ actions.
				\end{tabular} & & \\
				
				\hline
				\multirow{4}{*}[-3.5ex]{\begin{tabular}{l} 
						$Longest/Shortest$ \\
						$Action$
				\end{tabular}} & \textbf{longest\_action} & \multirow{2}{*}[-1.5ex]{ (exe\_trace, token) } & \multirow{2}{*}[-1.5ex]{ action } \\
				\cline{2-2} & \begin{tabular}{l} 
					Getting the action that takes the longest
					time \\ to occur.
				\end{tabular} & & \\
				\cline{2-4} & \textbf{shortest\_action} & 
				\multirow{2}{*}[-1.5ex]{ (exe\_trace, token) } & \multirow{2}{*}[-1.5ex]{ action } \\
				\cline{2-2} & \begin{tabular}{l} 
					Getting the action that takes the shortest
					time \\ to occur.
				\end{tabular} & & \\
				
				\hline
				\multirow{2}{*}[-1.5ex]{$Action$} & \textbf{filter\_actions} & \multirow{2}{*}[-1.5ex]{ (exe\_trace, token) } & \multirow{2}{*}[-1.5ex]{ actions } \\
				\cline{2-2} & \begin{tabular}{l} 
					Selects actions from the input scene
					list with \\the input relation ``doing”.
				\end{tabular} & & \\
				
				\hline
				\multirow{2}{*}[-1.5ex]{$Object$} & \textbf{query\_subject\_relation} & \multirow{2}{*}[-1.5ex]{ (exe\_trace, token) } & \multirow{2}{*}[-1.5ex]{ object } \\
				\cline{2-2} & \begin{tabular}{l} 
					Selects objects that involved with the
					specified \\ interaction.
				\end{tabular} & & \\
				\hline  
			\end{tabular}
		}
	\end{center}
\end{table*}

\begin{table*}
	\centering
	\caption{Reasoning rules corresponding to $Choose$, $Equals$, $Conjunction$, and $First/last$ question.}
	\label{tab14_rules_difficult}
	\begin{center}
		\scalebox{1}{
			\begin{tabular}{|c|c|c|c|}
				\hline \begin{tabular}{c} 
					Sub-question Type
				\end{tabular} & Reasoning Rule/ Description & Input Type & Output Type \\
				\hline
				\multirow{8}{*}[-5.5ex]{$Choose$} & \textbf{choose} & \multirow{2}{*}[-1.5ex]{ (exe\_trace, token) } & \multirow{2}{*}[-1.5ex]{ object } \\
				\cline{2-2} & \begin{tabular}{l} 
					Choose an object of two possible options,\\
					such as: ``equals”.
				\end{tabular} & & \\
				\cline{2-4} & \textbf{or} & \multirow{2}{*}[-1.5ex]{ (exe\_trace, token) } & \multirow{2}{*}[-1.5ex]{ time } \\
				\cline{2-2} & \begin{tabular}{l} 
					Choose a word of two possible options,\\
					such as: ``after” and ``before".
				\end{tabular} & & \\
				\cline{2-4} & \textbf{choose\_action\_shorter} & \multirow{2}{*}[-1.5ex]{ (exe\_trace, token) } & \multirow{2}{*}[-1.5ex]{ action } \\
				\cline{2-2} & \begin{tabular}{l} 
					Choose a shorter action from two possible\\
					action.
				\end{tabular} & & \\
				\cline{2-4} & \textbf{choose\_action\_longer} & \multirow{2}{*}[-1.5ex]{ (exe\_trace, token) } & \multirow{2}{*}[-1.5ex]{ action } \\
				\cline{2-2} & \begin{tabular}{l} 
					Choose a longer action from two possible\\
					action.
				\end{tabular} & & \\
				
				\hline
				\multirow{4}{*}[-2.5ex]{$Equals$} & \textbf{object\_equals} & \multirow{2}{*}[-1.5ex]{ (exe\_trace, token) } & \multirow{2}{*}[-1.5ex]{ binary } \\
				\cline{2-2} & \begin{tabular}{l} 
					Compares two objects to verify if they are\\
					the same.
				\end{tabular} & & \\
				\cline{2-4} & \textbf{action\_equals} & 
				\multirow{2}{*}[-1.5ex]{ (exe\_trace, token) } & \multirow{2}{*}[-1.5ex]{ binary } \\
				\cline{2-2} & \begin{tabular}{l} 
					Compares two actions to verify if they are\\
					the same.
				\end{tabular} & & \\
				
				\hline
				\multirow{4}{*}[-2.5ex]{$Conjunction$} & \textbf{conjunction\_and} & \multirow{2}{*}[-1.5ex]{ (exe\_trace, token) } & \multirow{2}{*}[-1.5ex]{ binary } \\
				\cline{2-2} & \begin{tabular}{l} 
					Combine two interaction questions using a\\
					conjunction ``and”.
				\end{tabular} & & \\
				\cline{2-4} & \textbf{conjunction\_xor} & 
				\multirow{2}{*}[-1.5ex]{ (exe\_trace, token) } & \multirow{2}{*}[-1.5ex]{ binary } \\
				\cline{2-2} & \begin{tabular}{l} 
					Combine two interaction questions using a\\
					conjunction ``xor”.
				\end{tabular} & & \\
				
				\hline
				\multirow{4}{*}[-2.5ex]{$First/last$} & \textbf{query\_first} & \multirow{2}{*}[-1.5ex]{ (exe\_trace, token) } & \multirow{2}{*}[-1.5ex]{ object/action } \\
				\cline{2-2} & \begin{tabular}{l} 
					Getting the first instance of the given
					object \\or action.
				\end{tabular} & & \\
				\cline{2-4} & \textbf{query\_last} & 
				\multirow{2}{*}[-1.5ex]{ (exe\_trace, token) } & \multirow{2}{*}[-1.5ex]{ object/action } \\
				\cline{2-2} & \begin{tabular}{l} 
					Getting the last instance of the given
					object\\ or action.
				\end{tabular} & & \\
				
				\hline  
			\end{tabular}
		}
	\end{center}
\end{table*}

\begin{table*}
	\centering
	\caption{Input/output data types of reasoning rules in the program executor.}
	\label{tab15_rules_in/out}
	\begin{center}
		\scalebox{1}{
			\begin{tabular}{|l|c|}
				\hline Type & Semantics \\
				\hline scene & \begin{tabular}{c} 
				The structural static and dynamic scene representation: person, objects, relations, actions, \\
				and its starting and ending time.
				\end{tabular} \\
				\hline exe\_trace & A dictionary that tracks the intermediate result of previous reasoning steps. \\
				\hline token & A string representing the name of the symbolic function to be executed currently. \\
				\hline binary & A string indicating a binary question out of ``yes", ``no". \\
				\hline time & A string indicating a temporal relation out of ``after", ``before". \\
				\hline object & \begin{tabular}{c} 
					A string indicating an object out of ``chair", ``paper", ``food", ``door",``vacuum", ``person", \\ ``laptop", ``dish", ``phone", ``blanket", ``doorknob" ``clothes", ``window", ``bed", ``floor",\\``closet", ``broom", ``mirror", ``table", ``refrigerator", ``pillow", ``picture",``bag", ``box", \\``light", ``shoe", ``medicine", ``doorway", 
					 ``television".
				\end{tabular} \\
				\hline objects & A list of the specified object in the scene. \\
				\hline relation & \begin{tabular}{c} 
					A string indicating a relation out of ``looking at", ``not looking at", ``unsure", ``above", \\
					``beneath", ``in front of", ``behind", ``on the side of", ``in", ``carrying", ``covered by", \\
					 ``drinking from",``eating", ``having it on the back", ``holding", ``leaning on", ``lying on",\\ 
					``sitting on", ``standing on", ``touching", ``twisting", ``wearing", ``wiping", ``writing on",\\
					``not contacting",``drinking", ``putting", ``taking", ``closing", ``throwing", ``putting down",\\ 
					``grasping", ``walking",``sitting”, ``watching”, ``opening”, ``snuggling”, ``standing”, ``working\\
					on”, ``tidying”, ``working”, ``awakening”, ``fixing”, ``smiling”, ``playing”, ``lying”, ``playing\\
					on”, ``sneezing”, ``dressing”, ``undressing”, ``washing”, ``pouring”, ``turning”, ``making”,\\ 
					``going”, ``talking”, ``consuming”, ``laughing”, ``running”, ``reaching”, ``photographing”,\\
					``cooking”.
				\end{tabular} \\
				\hline relations & A list of the specified relation in the scene. \\
				\hline action & \begin{tabular}{c} 
					A string indicating an action out of ``undressing themselves”, ``fixing a vacuum”,\\
					``washing a mirror”, ``holding a bag”, ``snuggling with a pillow”, ``watching a picture”, \\
					``watching a laptop	or something on a laptop”, ``fixing a door”, ``holding a vacuum”, \\
					``putting on a shoe”,``holding some food”, ``washing something with a blanket”, ``watching\\
					a book”, ``turning off a light”, ``holding a blanket”, ``watching television”, ``holding a mirror”,\\
					``taking off some shoes”, ``sitting at a table”, ``washing a window”, ``fixing their hair”, ``fixing \\
					a doorknob”, ``tidying up a blanket”, ``holding a book”, ``washing a cup”, ``lying on the floor”, \\
					``tidying up with a broom”, ``holding a paper”, ``smiling at something”, ``working on a book”, \\
					``holding a broom”, ``holding a cup of something”, ``watching something in a mirror”, \\
					``holding some medicine”, ``laughing at  something”, ``fixing a light”, ``snuggling with \\
					a blanket”, ``holding some clothes”, ``holding a phone”, ``washing some clothes”, ``holding\\
					a picture”, ``pouring something into a cup”, ``dressing	themselves”, ``tidying up a closet”,\\ ``sitting in a bed”, ``holding a shoe”, ``holding a pillow”, ``washing their hands”, ``None”,\\ ``turning on a light”, ``lying on a bed”, ``tidying some clothes”, ``washing a table”, ``tidying\\ something on the floor”, ``sitting on the floor”, ``tidying up a table”, ``standing up”, ``walking\\ through a doorway”, ``eating some food”, ``holding a dish”, ``standing on a chair”, ``watching\\ outside of a window”, ``grasping onto a doorknob”, ``holding a box”,``running somewhere”, \\
					``sitting in a chair”, ``holding a laptop”, ``making some food”, ``sitting on a table”, ``awakening\\ in bed”, ``sneezing somewhere”.
				\end{tabular} \\
				\hline actions & A list of the specified relation ``doing” in the scene. \\
				\hline
			\end{tabular}
		}
	\end{center}
\end{table*}


\end{document}